\documentclass[acmtog,nonacm]{acmart}

\setcopyright{none}
\makeatletter
\def\@mkbibcitation{\bgroup
  \let\@vspace\@vspace@orig
  \let\@vspacer\@vspacer@orig
  \def\footnotemark{}%
  \def\\{\unskip{} \ignorespaces}%
  \par\medskip\small\noindent{\bfseries ACM Reference Format:}\par\nobreak
  \noindent\bgroup\def\\{\unskip{}, \ignorespaces}\authors\egroup. 2026. \@title.
  \textit{ACM Trans. Graph.} 45, 6, Article~265 (December 2026).
  \url{https://doi.org/10.1145/3842583}
  \par\egroup}
\makeatother

\fancypagestyle{arxivstyle}{%
  \fancyhf{}
  \fancyfoot[C]{\small\thepage}
  \renewcommand{\headrulewidth}{0pt}
  \renewcommand{\footrulewidth}{0pt}
}

\usepackage{booktabs}
\usepackage{xcolor}

\usepackage[linesnumbered,ruled,vlined]{algorithm2e}

\SetAlFnt{\small}
\SetAlCapFnt{\small}
\SetAlCapNameFnt{\small}
\SetAlCapHSkip{0pt}

\usepackage{xcolor}
\usepackage{enumitem}
\usepackage{xspace}
\usepackage{booktabs}
\usepackage{colortbl}
\usepackage{multirow}
\usepackage{makecell}
\usepackage{lipsum}
\usepackage{graphicx}
\usepackage{titletoc}
\usepackage{amsmath,amsfonts,bm}
\usepackage{wrapfig}

\definecolor{DarkRed}{RGB}{139, 0, 0}

\usepackage[normalem]{ulem}
\usepackage[labelsep=period]{caption}

\definecolor{MyDarkRed}{rgb}{0.66, 0.16, 0.16}
\definecolor{MyDarkBlue}{rgb}{0.16, 0.16, 0.66}

\definecolor{Red}{rgb}{0.6,0,0}
\definecolor{Blue}{rgb}{0,0,0.8}
\definecolor{Green}{rgb}{0.2,0.8,0}
\definecolor{airforceblue}{rgb}{0.36, 0.54, 0.66}
\definecolor{ao(english)}{rgb}{0.0, 0.5, 0.0}
\definecolor{azure(colorwheel)}{rgb}{0.0, 0.5, 1.0}
\definecolor{crimson}{rgb}{0.86, 0.08, 0.24}
\definecolor{darkcerulean}{rgb}{0.03, 0.27, 0.49}
\definecolor{cobalt}{rgb}{0.0, 0.28, 0.67}
\definecolor{rosegold}{rgb}{0.72, 0.43, 0.47}
\definecolor{orange-red}{rgb}{1.0, 0.27, 0.0}
\definecolor{mountainmeadow}{rgb}{0.19, 0.73, 0.56}
\definecolor{malachite}{rgb}{0.04, 0.85, 0.32}
\definecolor{darkblue}{rgb}{0.0, 0.0, 0.55}
\definecolor{customblue}{rgb}{0.2, 0.35, 0.8}

\newcommand{\hidden}{\mathbf{h}}

\newcommand{\token}{\mathbf{T}}

\makeatletter
\def\@titlefont{\fontsize{14.6}{18}\selectfont\sffamily}
\newcommand{\methodmark}[1]{#1}
\newcommand{\titlemethodmark}[1]{{\bfseries\color{magenta}#1}}
\def\@mktitle@i{\hsize=\textwidth
  \@ACM@title@width=\hsize
  \setbox\mktitle@bx=\vbox{%
    \noindent{\color{magenta}\rule{28pt}{2.4pt}}\par\vskip 6pt
    \noindent\@titlefont
    \parbox[t]{\@ACM@title@width}{\raggedright\@titlefont\noindent
      \let\methodmark\titlemethodmark
      \@title\@translatedtitle
      \ifx\@subtitle\@empty\else
        \par\noindent{\@subtitlefont\@subtitle\@translatedsubtitle}%
      \fi}%
    \par\bigskip}}
\pretocmd{\@mkteasers}{%
  \global\setbox\mktitle@bx=\vbox{\unvbox\mktitle@bx\par }}{}{}
\def\@mkabstract{\bgroup
  \ifx\@abstract\@lempty\else
  {\phantomsection\addcontentsline{toc}{section}{\abstractname}%
   \noindent{\@secfont\headingstyle\abstractname}\par\vskip .25\baselineskip
   \small
   \everypar{\setbox\z@\lastbox\everypar{}}%
   \ignorespaces\@abstract\par}%
  \fi\egroup}
\AtBeginDocument{\pdfstringdefDisableCommands{\let\methodmark\@firstofone}}
\makeatother

\usepackage{pdfrender}
\definecolor{headingcolour}{HTML}{29426A}
\newcommand{\headingstyle}{\color{headingcolour}\pdfrender{TextRenderingMode=FillStroke,LineWidth=0.18pt}}
\makeatletter
\patchcmd{\NR@sect}{\@hangfrom{\hskip #3\relax\@svsec}}
  {\@hangfrom{\hskip #3\relax{\headingstyle\@svsec}}}{}{\errmessage{patch 1 failed}}
\patchcmd{\NR@sect}{\interlinepenalty \@M #8\@@par}
  {\interlinepenalty \@M {\headingstyle #8}\@@par}{}{\errmessage{patch 2 failed}}
\patchcmd{\NR@ssect}{\interlinepenalty \@M #5\@@par}
  {\interlinepenalty \@M {\headingstyle #5}\@@par}{}{\errmessage{patch 3 failed}}
\makeatother
\usepackage{tikz}
\tikzset{teaserlabel/.style={anchor=south west, inner xsep=2.8pt,
  inner ysep=1.4pt, text height=4.4pt, text depth=1.3pt, text=white,
  font=\sffamily\fontsize{6}{7}\selectfont}}
\newcommand{\teasertab}[3]{%
  \node[teaserlabel, text opacity=0] (tt) at (#1,#2) {#3};
  \fill[black, fill opacity=0.55] (tt.south west) -- (tt.north west)
    -- ([xshift=-2pt]tt.north east) arc[start angle=90, end angle=0, radius=2pt]
    -- (tt.south east) -- cycle;
  \node[teaserlabel] at (#1,#2) {#3};}
\newcommand{\teaserlabels}{%
  \teasertab{0.7504}{0.8002}{Albedo}
  \teasertab{0.7504}{0.6004}{Metallic}
  \teasertab{0.7504}{0.4003}{Roughness}
  \teasertab{0.5003}{0.0000}{Close-up}}
\definecolor{citecolour}{HTML}{334D76}
\AtEndPreamble{\hypersetup{colorlinks=true,citecolor=citecolour,
  linkcolor=black,urlcolor=black,filecolor=black}}
\begin{document}
\fancypagestyle{firstpagestyle}{%
  \fancyhf{}%
  \renewcommand{\headrulewidth}{0pt}\renewcommand{\footrulewidth}{0pt}%
  \fancyfoot[C]{\small\thepage}}
\settopmatter{printacmref=true}
\newcommand{\method}{UltraTex}
\title{\methodmark{\method:} Aligning Physical Knowledge for High-Fidelity PBR Texture Generation}

\author{Jingzhi Bao}
\orcid{0009-0006-8748-0511}
\affiliation{%
  \institution{SSE, CUHKSZ}
  \city{Shenzhen}
  \country{China}
}
\email{jingzhibao03@gmail.com}

\author{Xiaohe Ma}
\orcid{0009-0000-8924-2046}
\affiliation{%
  \institution{Meshy AI}
  \city{Shanghai}
  \country{China}
}
\email{xiaohema@meshy.ai}

\author{Hong Li}
\orcid{0000-0002-4240-3073}
\affiliation{%
  \institution{Meshy AI}
  \city{Shanghai}
  \country{China}
}
\email{hongli@meshy.ai}

\author{Yuanming Hu}
\orcid{0000-0003-3791-0705}
\affiliation{%
  \institution{Meshy AI}
  \city{Shanghai}
  \country{China}
}
\email{yuanmhu@gmail.com}

\author{Xiaoguang Han}
\orcid{0000-0003-0162-3296}
\authornote{Corresponding author.}
\affiliation{%
  \institution{SSE, CUHKSZ and SLAI}
  \city{Shenzhen}
  \country{China}
}
\email{hanxiaoguang@cuhk.edu.cn}

\authorsaddresses{}

\begin{abstract}

Generating physically based rendering (PBR) texture maps for a 3D mesh from a single reference image is highly ambiguous, as the observed appearance entangles material identity, surface reflectance, and illumination.
While current generative methods produce realistic albedo maps, they still struggle to infer accurate metallic and roughness attributes for relighting.
We present \method{}, a cascaded multi-view framework that moves beyond appearance matching by integrating physical knowledge into multi-view diffusion-based PBR generation.
\method{} first jointly predicts albedo and world-space shading normals to generate more detailed albedo maps, and then predicts metallic-roughness (MR) maps conditioned on the generated albedo, reference image, and material semantic priors.
To inject material semantics, a lightweight adapter transforms hidden-layer features of a vision-language model (VLM) into intrinsic material representation tokens, providing part-level knowledge anchors for MR estimation.
To model material-light interaction, photometric representation alignment guides intermediate diffusion features toward ground-truth material renderings under controlled lighting.
This makes metallic and roughness variations observable through highlight intensity, specular sharpness, and local reflectance contrast, providing stronger guidance for fine-grained material prediction without explicit lighting supervision.
Extensive experiments show that \method{} generates high-quality textures and outperforms existing open-source and commercial methods in visual fidelity, material accuracy, and physical plausibility.

\end{abstract}



\begin{teaserfigure}
  \begin{tikzpicture}
    \node[anchor=south west, inner sep=0pt, outer sep=0pt] (T) at (0,0)
      {\includegraphics[width=\textwidth]{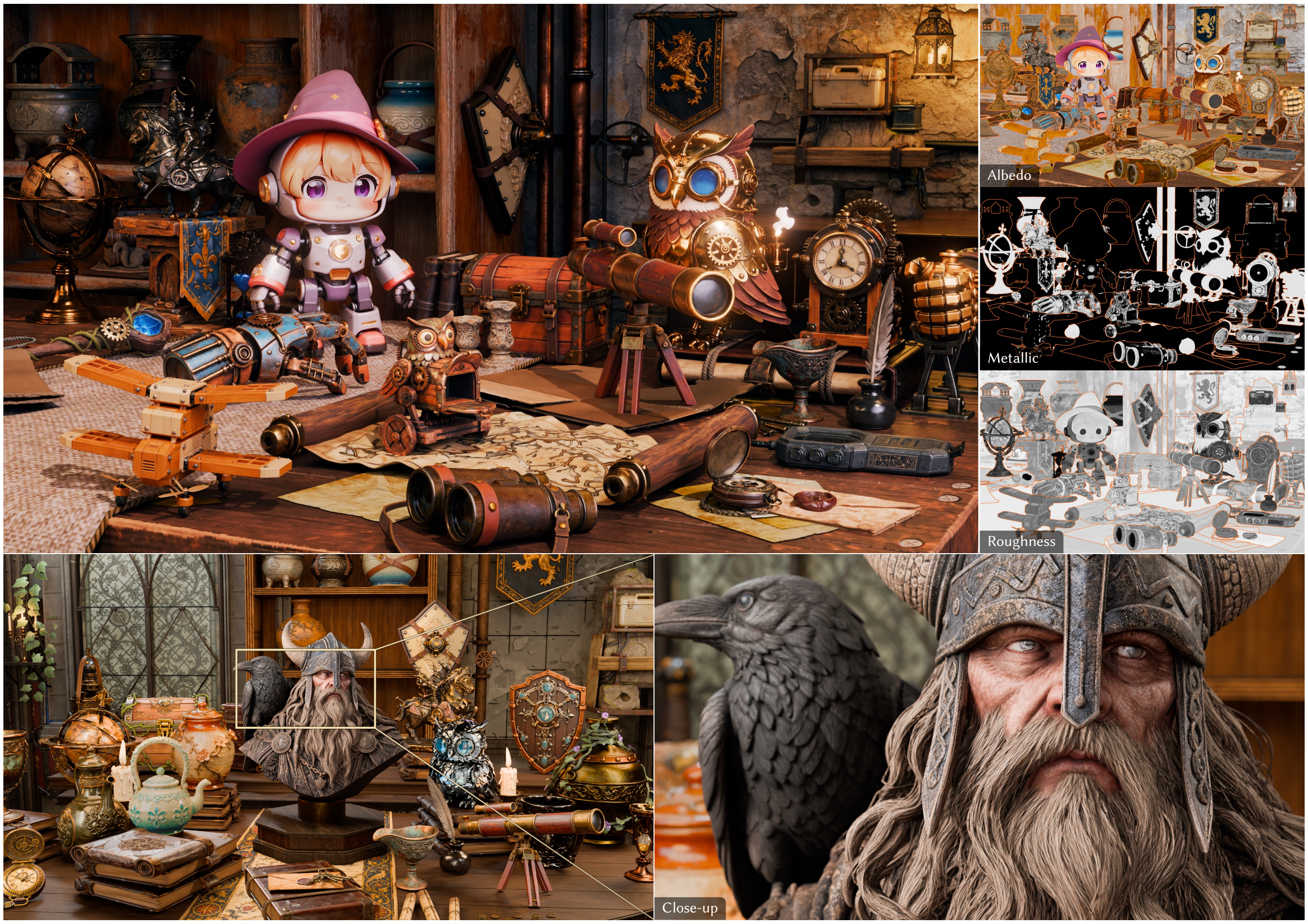}};
    \begin{scope}[x={(T.south east)}, y={(T.north west)}]
      \teaserlabels
    \end{scope}
  \end{tikzpicture}
  \vspace{-2em}
  \caption{\textbf{PBR texture gallery generated by \method.}
    Our method produces detailed and physically plausible PBR materials across diverse assets under environmental lighting.
    In the first row, the right column shows the generated albedo, metallic, and roughness maps from top to bottom.
    In the second row, a close-up render (right) of the boxed region (left) reveals fine-grained material details.
    Project page: \href{https://ultratex.vercel.app/}{\textcolor{magenta}{\nolinkurl{https://ultratex.vercel.app/}}}.}
  \label{fig:teaser}
\end{teaserfigure}

\maketitle
\pagestyle{arxivstyle}

\section{Introduction}

Physically based rendering (PBR) is the industry standard for material representation of 3D assets in games, films, and AR/VR~\citep{burley2012disney,PBRT3e}.
PBR texture maps separate visible appearance from intrinsic material attributes, enabling assets to respond plausibly under different lighting conditions.
While albedo defines the base color of a surface, realistic relighting depends critically on properties such as metallic and roughness.
These attributes describe physical reflectance behavior rather than directly observable image signals, making high-quality PBR texture authoring a labor-intensive process that still relies heavily on expert-driven workflows~\cite{substance3dpainter}.

Generating PBR texture maps for a 3D mesh from a single reference image is therefore highly ambiguous.
The observed appearance entangles material identity, surface reflectance, geometry, and illumination.
For example, a bright region in the input may arise from polished metal, glossy dielectric coating, baked illumination, or a local color pattern.
This ambiguity is especially severe for metallic--roughness (MR) estimation, where the desired outputs are abstract intrinsic maps rather than natural RGB images.

Early inverse-rendering methods reduce this ambiguity by using controlled capture setups, such as co-located flash lighting, to expose physical cues like specular highlights~\cite{guo2021highlight,ma2021scanner,bao2025pir}.
Although effective in constrained environments, these methods do not generalize well to arbitrary in-the-wild images.
Recent generative approaches instead leverage powerful image-domain priors from diffusion models~\citep{rombach2022high,esser2024sd3,blackforest2024flux} and multi-view generation techniques~\citep{shi2023mvdream,huang2024mvadapter,hunyuan3d2025hunyuan3d2.1,SEED3D} to synthesize multi-view texture or PBR images, which are then reprojected onto the mesh surface to construct texture maps.
While this paradigm has substantially improved albedo generation, it remains unreliable for intrinsic MR prediction.
Existing methods often assign metallic values to incorrect regions or produce overly conservative roughness maps with limited fine-scale variation.
These failures indicate that strong appearance priors alone are insufficient: the model must also understand what each object part is made of and how its material should respond to light.

Professional PBR creation follows this reasoning naturally: artists infer material maps by considering asset style, part-wise material composition, surface structure, and illumination behavior, rather than matching pixels alone.
We argue that high-fidelity PBR generation requires aligning such physical knowledge with the generative process.
This alignment can be achieved in two complementary ways.
First, physical knowledge can be injected as conditioning priors for material semantics.
Vision-language models are well suited for this role, as their hidden representations encode rich world knowledge and semantic reasoning from large-scale image-text pretraining; however, an adapter is needed to translate them into diffusion-compatible material tokens.
Second, physical knowledge can guide learning through training objectives.
Since metallic and roughness maps are abstract intrinsic parameters whose effects become visible only after interacting with albedo, geometry, and illumination, MR generation should be supervised in a photometric representation space where controlled-light renderings reveal material-light responses.

To this end, we introduce \method{}, a cascaded multi-view framework for high-quality PBR texture generation from a single reference image and an input mesh.
\method{} first jointly predicts albedo and world-space shading normals to obtain detailed albedo maps, and then predicts MR maps conditioned on the generated albedo, reference-image features, and material semantic priors.
For material-semantic conditioning, we introduce a VLM-guided intrinsic material adapter that transforms hidden-layer vision-language features into intrinsic material tokens, providing part-level knowledge anchors for MR estimation.
For photometric guidance, we propose photometric representation alignment, which guides intermediate MR diffusion features toward visual representations of ground-truth materials rendered under controlled lighting.
This rendered RGB space reveals fine-grained MR variations through highlight intensity, specular sharpness, and local reflectance contrast, providing stronger guidance without explicit lighting supervision.

Extensive experiments demonstrate that \method{} generates high-quality PBR textures and substantially outperforms state-of-the-art open-source and commercial methods in visual fidelity, material accuracy, and physical plausibility.

In summary, we make the following contributions:
\begin{itemize}[topsep=2pt, leftmargin=12pt]
    \item To the best of our knowledge, we present the first diffusion-based PBR texture generation framework that explicitly aligns physical knowledge through material-semantic conditioning and photometric representation guidance.
    \item We introduce a VLM-guided intrinsic material adapter that translates hidden-layer vision-language features into material-aware tokens, injecting part-wise semantic priors into MR generation.
    \item We propose photometric representation alignment, which guides intermediate MR diffusion features with controlled-light material renderings to capture fine-grained material-light interactions.
    \item Extensive experiments demonstrate that \method{} substantially outperforms existing open-source and commercial methods in visual fidelity, material accuracy, and physical plausibility.
\end{itemize}

\section{Related Work}

\subsection{Texture Generation}

Foundation models have opened new directions for automating texture generation in 3D content creation.
Early methods adopt optimization-based paradigms, using Score Distillation Sampling (SDS)~\cite{poole2023dreamfusion} to distill priors from pretrained 2D diffusion models into 3D assets~\citep{poole2022dreamfusion,lin2023magic3d,wang2023prolificdreamer,po2024compositional,metzer2022latent,chen2023scenetex,khalid2022clipmesh,text2mesh,chen2022tango}.
However, iterative optimization over renderings is costly and often produces over-saturated textures or artifacts that limit practical use.

Feed-forward approaches explored progressive inpainting across pre-defined views conditioned on explicit 3D attributes~\citep{yu2023texture,bensadoun2024meta,cao2023texfusion,liu2024text}.
However, the geometric fidelity is frequently compromised by the cross-view latent synchronization. Alternatively, to bypass such occlusion and alignment issues, other methods directly target the UV space~\citep{yu2023texture,zeng2024paint3d,bensadoun2024meta,yu2024texgen}.
Yet, the topological fragmentation inherent to UV parameterizations severely degrades the generative capacity of 2D diffusion models, hindering high-fidelity synthesis.

Recent methods
{~\citep{huang2024mvadapter,zhao2025hunyuan3d,liu2025calitex,yeo2026matlat,he2026hitem3d}
}
incorporate multi-view attention mechanisms~\citep{li2024era3d,kant2024spad,huang2024epidiff,shi2023mvdream,wang2023imagedream,feng2025romantex} to generate globally consistent textures while respecting 3D geometry.
However, they mainly target appearance texturing, producing either albedo maps or illumination-baked textures without intrinsic material attributes required for physically based relighting.

{
Another line of work performs texture generation directly in native 3D space~\citep{lai2025natexseamlesstexturegeneration,chen2026lafite,textrix2025,xiang2025trellis2}. These methods directly generate complete surface textures in 3D space, providing inherent geometric consistency while avoiding view-projection artifacts such as edge bleeding. However, the scarcity of high-quality 3D data with high-resolution textures limits fine-detail modeling.
}

\begin{figure}[t]
    \centering
    \includegraphics[width=\linewidth]{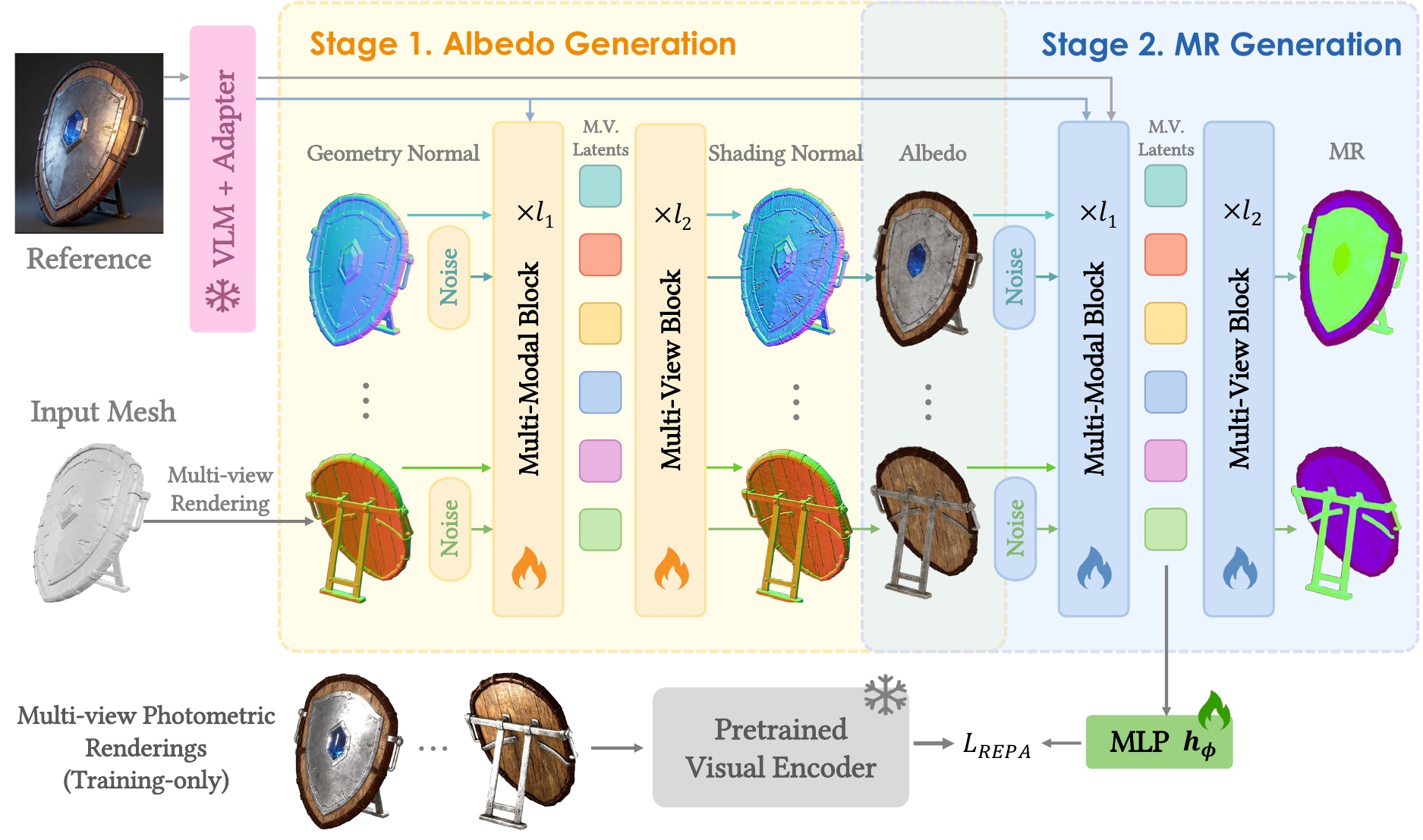}
    \vspace{-0.5em}
    \vspace{-.5em}
    \caption{\textbf{Overall pipeline of \method.}
    \method{} adopts a cascaded two-stage multi-view diffusion pipeline.
    Stage 1 jointly predicts albedo maps and world-space shading normals from the reference image and multi-view geometry normals.
    Stage 2 predicts metallic-roughness (MR) maps (where purple and green represent roughness and metallic, respectively) conditioned on the generated albedo, reference-image features, and VLM-derived intrinsic material tokens.
    For MR training, photometric representation alignment uses a frozen visual encoder to extract representations from controlled-light material renderings, and an MLP aligns intermediate diffusion features to this photometric representation space via $\mathcal{L}_{\mathrm{REPA}}$.}
    \label{fig:method}
    \vspace{-2em}
\end{figure}

\subsection{PBR Texture Generation}
Building upon texture synthesis, recent methods extend pretrained diffusion models to PBR material generation.
Early works incorporate BRDF formulations into SDS optimization to infer material properties~\citep{chen2023fantasia3d,zhang2024dreammat,liu2024unidream,youwang2024paint,wu2023hyperdreamer,xu2023matlaber,yeh2024texturedreamer}, but often struggle to decompose lighting and surface reflectance, leading to unrealistic or flat results.

To enhance spatial alignment and material accuracy, subsequent approaches either employ iterative latent synchronization~\citep{huang2024materialanything,zhang2024clay} or leverage LLMs for retrieval-augmented generation~\citep{zhang2024mapa,dang2024texpro,wang2024boosting,fang2024make,wei2025pbr3dgen}.
Despite this, these methods primarily transfer generative priors to the PBR task without end-to-end training on large-scale PBR datasets, which fundamentally constrains their generalization ability.

A recent line of work fine-tunes image or video models to predict multi-channel PBR maps
{
~\citep{he2025materialmvp,zhu2024mcmat,vainer2024collaborative,vecchio2024matfuse,hasselgren2026videomatgen,ma2025materialpicker,munkberg2025videomat,yuan2025seqtex}.
These methods mainly transfer appearance-generation priors to intrinsic material prediction through pixel-level reconstruction losses.
While scalable, this transfer does not directly resolve the ambiguity of MR estimation, where metallic and roughness are latent material properties rather than observable appearance.
As a result, these models remain biased toward appearance matching, leading to frequent ambiguity in intrinsic material decomposition.
In contrast, our method introduces explicit material reasoning from VLM world knowledge and representation-level photometric regularization to provide semantic and physical supervision for MR generation.
}

\section{Method}

Given a single reference image and an input mesh, \method{} generates cross-view consistent PBR texture maps through a cascaded two-stage multi-view diffusion pipeline (Fig.~\ref{fig:method}).
Stage 1 jointly predicts orthographic-view albedo maps and world-space shading normals from the reference image and multi-view geometry normals (Sec.~\ref{sec:albedo}).
Conditioned on the generated albedo, Stage 2 predicts metallic-roughness (MR) maps with a VLM-guided intrinsic material adapter (Sec.~\ref{sec:vlm}) and photometric representation alignment (Sec.~\ref{sec:photometric}).
Both stages are built upon an image diffusion backbone and are trained separately; following~\cite{hunyuan3d2025hunyuan3d2.1}, MR maps are represented as two channels of a single RGB image.

\begin{figure}
    \centering
    \includegraphics[width=\linewidth]{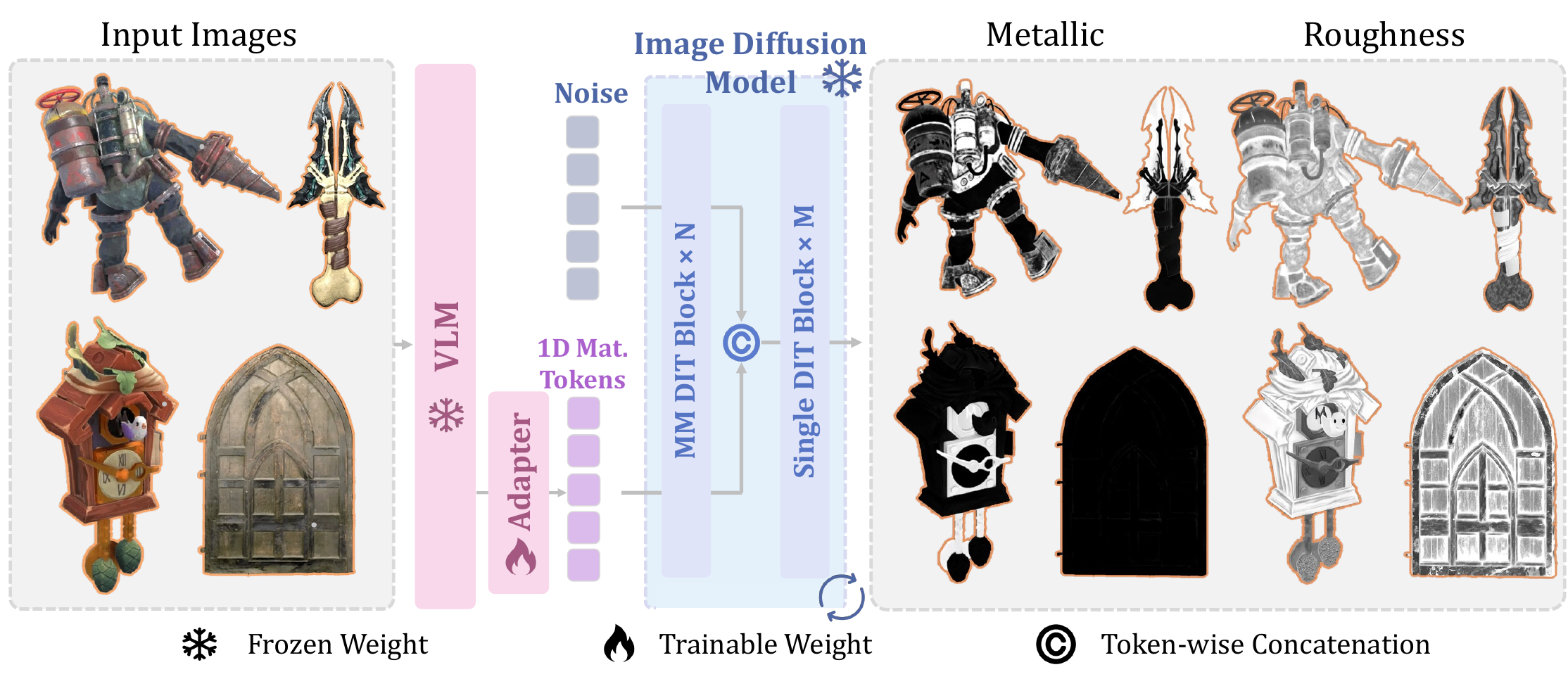}
    \vspace{-1.5em}
    \caption{\textbf{VLM-guided intrinsic material adapter.}
    A frozen VLM extracts hidden-layer visual features from rendered reference images, and a trainable adapter projects them into 1D intrinsic material tokens.
    With the image diffusion model frozen, the adapter is trained to reconstruct ground-truth metallic and roughness maps by injecting these tokens into the denoising blocks.
    The learned tokens encode part-aware material semantic priors and are used to guide MR generation.}
    \vspace{-1em}
    \label{fig:vlm_res}
\end{figure}

\subsection{Multi-view Albedo and Shading Normal Generation}
\label{sec:albedo}

The first stage generates multi-view albedo maps and world-space shading normals from the reference image and rendered geometry normals. We adopt a two-step architecture: a Multi-Modal Transformer (MM-T) fuses conditioning tokens and noisy view latents per view, then a Multi-View Transformer (MV-T) models cross-view interactions for global consistency.

\subsubsection{Multi-Modal DiT}
\label{sec:mm-dit}
For each view, MM-T integrates the ref\-er\-ence-image and geometry-normal conditions into a unified latent representation.
Specifically, for the $i$-th view, we extract geometry-normal tokens $\token_{\mathrm{geo}}^i$ by encoding the rendered geometry normal $\bm{G}_i$ with a fixed VAE encoder $\mathcal{E}_{\mathrm{VAE}}$, and derive reference-image tokens $\token_{\mathrm{img}}$ from both VAE latents and DINOv2 embeddings.
These conditional tokens, together with the noisy view latent $\bm{z}_i$, are processed by $l_1$ transformer blocks with intra-view self-attention:
\begin{align}
\token_{\mathrm{geo}}^i &=
        \mathrm{Linear}_{\mathrm{geo}}
        \left[
            \mathcal{E}_{\mathrm{VAE}}(\bm{G}_i)
        \right], \label{eq:token_geo} \\
\token_{\mathrm{img}} &=
        \mathrm{Linear}_{\mathrm{img}}
        \left[
            \mathcal{E}_{\mathrm{VAE}}(\bm{I}) ; \operatorname{DINOv2}(\bm{I})
        \right], \label{eq:token_img} \\
\bm{z}_i^{l_1} &=
        \operatorname{MM\text{-}T}^{l_1}
        \left(
            \bm{z}_i, \token_{\mathrm{img}}, \token_{\mathrm{geo}}^i
        \right). \label{eq:z_fusion}
\end{align}

\subsubsection{Multi-View DiT}
\label{sec:mv-dit}

After per-view multi-modal fusion, we perform global reasoning across views. To focus this stage strictly on cross-view interaction, we discard all conditional tokens ($\token_{\text{img}}$, $\token_{\text{geo}}$) and retain only the view-wise latents $\bm{z}_i$ from all $N$ views. Formally, the intermediate features from all views are aggregated into a unified sequence and processed through a series of $l_2$ Transformer blocks. In this joint representation $\{\hat{z}_i\} \in \mathbb{R}^{NL \times C}$, self-attention is performed globally to capture cross-view correlations and ensure 3D coherence.
\begin{equation}
    \{\hat{\bm{z}}_i\}_{i=1}^{N}
    =
    \operatorname{MV\text{-}T}^{l_2}
    \left(
        \bm{z}_1^{l_1}, \bm{z}_2^{l_1}, \dots, \bm{z}_N^{l_1}
    \right)
    \in \mathbb{R}^{NL \times C}.
\end{equation}

The stage is trained with a flow matching objective, where $z_0$ denotes the noise-free latent of the ground-truth albedo map and $\mathbf{c}=\{\token_{\mathrm{img}}, \token_{\mathrm{geo}}\}$ denotes the conditioning tokens:
\begin{equation}
\mathcal{L}_{\mathrm{alb}}(\theta) =
\mathbb{E}_{t,z_0,\epsilon}
\left[
\left\|
v_\theta(z_t;t,\mathbf{c}) - (\epsilon - z_0)
\right\|^2
\right].
\label{eq:flow_matching}
\end{equation}

\begin{figure}
    \centering
    \includegraphics[width=\linewidth]{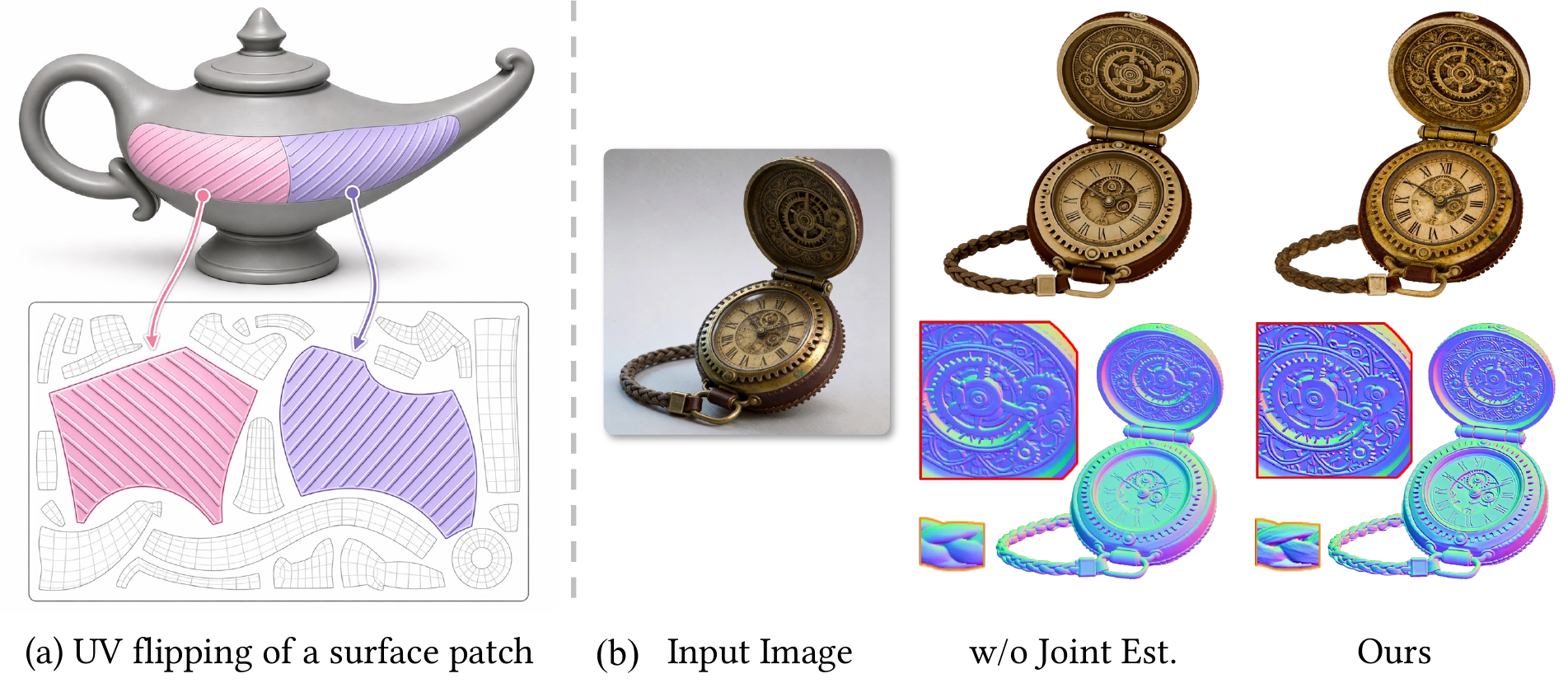}
    \vspace{-1.5em}
    \caption{\textbf{Effectiveness of joint albedo and shading-normal estimation.}
    (a) Tangent-space normal or bump maps depend on local UV parameterization; continuous 3D surfaces may be unwrapped into disjoint UV patches with different orientations, leading to inconsistent tangent frames.
    (b) By jointly estimating world-space shading normals with albedo, our model avoids such UV-dependent artifacts and produces richer surface details.
    We compare against a model trained without joint shading-normal estimation, where the normal visualization corresponds to the input geometry normals.}
    \vspace{-1em}
    \label{fig:bump_res}
\end{figure}

\subsubsection{Joint Estimation of Shading Normal}

We introduce a computationally efficient scheme to jointly predict \emph{world-space shading normals} ($\bm{S}$) alongside albedo maps.
This auxiliary prediction provides geometry-aware supervision for albedo generation, encouraging richer surface structure and high-frequency details.

\textit{Why world-space shading normals?}
Unlike tangent-space normal or bump maps, which depend on local UV coordinates, world-space shading normals are expressed in a global coordinate frame and remain independent of UV parameterization.
This avoids the inconsistent tangent frames introduced by rotated or flipped UV islands, making normal prediction less prone to seams (Fig.~\ref{fig:bump_res}(a)).

We denote the VAE latent of $\bm{S}$ as $\token_{\text{sn}}$.
Our key insight is that the geometry normal tokens $\token_{\text{geo}}$, which serve as input conditioning during albedo training, naturally reside in the same modality as world-space shading normals.
Leveraging this geometric alignment, we reuse the existing pathway to predict a joint velocity field.
Specifically, the model is optimized by a multi-task objective consisting of an albedo flow matching loss $\mathcal{L}_{\text{alb}}$ and a shading-normal refinement loss $\mathcal{L}_{\text{normal}}$.
While the albedo component follows the standard flow from Gaussian noise $\bm{\epsilon}$ to the ground-truth albedo latent $\token_{\text{alb}}$, the shading-normal component treats $\token_{\text{geo}}$ as the starting point to learn the residual micro-geometry:
\begin{align}
\mathcal{L}_{\text{alb}}(\theta) &= \mathbb{E}_{t, \token_{\text{alb}}, \bm{\epsilon}} \left[ \| v_{\theta, \text{alb}}(z_t; t, \mathbf{c}) - (\bm{\epsilon} - \token_{\text{alb}}) \|^2 \right], \label{eq:albedo_loss}\\
\mathcal{L}_{\text{normal}}(\theta) &= \mathbb{E}_{t, \token_{\text{sn}}, \token_{\text{geo}}} \left[ \| v_{\theta, \text{sn}}(z_t; t, \mathbf{c}) - (\token_{\text{geo}} - \token_{\text{sn}}) \|^2 \right], \\
\mathcal{L}_{\text{joint}} &= \mathcal{L}_{\text{alb}} + \lambda_{\text{normal}}\mathcal{L}_{\text{normal}},
\end{align}
where $\lambda_{\text{normal}}$ is a balancing weight, and $v_{\theta, \text{alb}}, v_{\theta, \text{sn}}$ represent the respective velocity components predicted by the model.

\subsection{Multi-view MR Generation}
Our multi-view MR generator follows the same MM-T/MV-T architecture as the albedo model in Sec.~\ref{sec:albedo}, but replaces the geometry-normal condition with the Stage-1 albedo maps.
It further introduces two physical knowledge alignment modules for intrinsic material estimation: a VLM-guided intrinsic material adapter and photometric representation alignment.

\subsubsection{VLM-Guided Intrinsic Material Adapter}
\label{sec:vlm}

Before training the multi-view MR generator, we train a VLM-guided intrinsic material adapter in a separate stage, as illustrated in Fig.~\ref{fig:vlm_res}.
The goal is to translate material-aware visual knowledge from a frozen vision-language model (VLM) into diffusion-compatible intrinsic material tokens.
During this adapter training stage, both the VLM and the image diffusion model are kept frozen, and only the adapter is optimized.

Previous studies~\cite{neotowards,kaduri2025vlmanalysis} indicate that VLM hidden representations evolve progressively across layers: early layers capture raw visual features, whereas intermediate layers encode more abstract semantic and reasoning cues.
Since MR generation requires reasoning about intrinsic material properties rather than direct appearance transfer, we exploit intermediate VLM tokens extracted from the encoded reference image $\bm{I}$ as a source of material-aware context.
Specifically, let $\mathcal{V}_m(\bm{I}) \in \mathbb{R}^{P \times d_{\text{vlm}}}$ denote the visual-token subsequence of the VLM hidden state at layer $m$, where $P$ is the number of image patch tokens.
We select a set of intermediate layers $\mathcal{M}$ and concatenate their visual tokens along the channel dimension:
\begin{equation}
\mathbf{H}^{\mathrm{vlm}}_{\mathcal{M}}
=
\operatorname{concat}_{m \in \mathcal{M}}
\left[
\mathcal{V}_{m}(\bm{I})
\right]
\in
\mathbb{R}^{P \times |\mathcal{M}|d_{\text{vlm}}}.
\end{equation}

The VLM token space is not directly compatible with the conditioning space of the image diffusion model.
We therefore introduce an \emph{Intrinsic Material Token Adapter} $\mathcal{A}$, which projects $\mathbf{H}^{\mathrm{vlm}}_{\mathcal{M}}$ into 1D intrinsic material tokens:
\begin{equation}
\token_{\text{mat}}
=
\mathcal{A}
\left(
\mathbf{H}^{\mathrm{vlm}}_{\mathcal{M}}
\right).
\end{equation}
These tokens are injected into the frozen image diffusion model through token-wise concatenation in the denoising blocks, where they interact with noisy latents to reconstruct ground-truth metallic and roughness maps.
We optimize the adapter with the standard flow-matching objective on rendered reference images paired with their ground-truth MR maps.
After training, the learned tokens encode part-aware material semantic priors and are used as conditional inputs for Stage-2 MR generation.

\subsubsection{Photometric Representation Alignment}
\label{sec:photometric}

Inspired by representation alignment methods such as REPA~\cite{yu2025repa}, we introduce photometric representation alignment to provide stronger guidance for MR generation.
Directly aligning raw MR maps with pretrained visual encoders is ill-posed, since encoders such as DINOv2~\cite{oquab2023dinov2} and MAE~\cite{he2022mae} are trained on natural images, while metallic and roughness maps are abstract intrinsic parameters.

To bridge this domain gap, we construct alignment targets in a photometrically meaningful RGB space.
For each training view, we render the ground-truth materials $\mathbf{x}$ into an auxiliary photometric response $\mathbf{x}_{\text{pm}}$ under a controlled point-light setup, where the light source is collocated with the camera.
The rendered response exposes material-light interactions, such as highlight intensity, specular sharpness, and local reflectance contrast, while remaining compatible with pretrained visual encoders.
We use a frozen DINOv2 encoder $f$ to extract the target representation $\mathbf{y} = f(\mathbf{x}_{\text{pm}})$.

During MR generation, the intermediate transformer features $\hidden_t$ at timestep $t$ are projected by a MLP layer $h_\phi$ and aligned with $\mathbf{y}$:
\begin{equation}
\mathcal{L}_{\mathrm{REPA}}(\theta,\phi)
=
-
\mathbb{E}_{(\mathbf{x},\mathbf{x}_{\text{pm}}),t}
\left[
\frac{1}{N_p}
\sum_{n=1}^{N_p}
\operatorname{sim}
\left(
\mathbf{y}_n,
h_\phi(\mathbf{h}_t^{[n]})
\right)
\right],
\label{eq:loss_repa}
\end{equation}
where $N_p$ denotes the number of patches and $\operatorname{sim}(\cdot,\cdot)$ is the patch-wise cosine similarity.
This representation-level alignment guides MR diffusion features toward the visual response of physically valid materials, providing stronger cues for fine-grained metallic and roughness prediction without explicit lighting supervision.

\subsubsection{Network}

The MR network follows the same MM-T and MV-T design as the albedo stage.
For the $i$-th view, we replace the geometry-normal condition with the generated albedo map $\bm{A}_i$ from Stage 1:
\begin{equation}
\token_{\text{alb}}^i =
\operatorname{Linear}_{\text{alb}}
\left[
\mathcal{E}_{\text{VAE}}(\bm{A}_i)
\right].
\end{equation}
Together with the reference-image tokens $\token_{\text{img}}$ and the VLM-derived material tokens $\token_{\text{mat}}$, the noisy MR latent is fused by MM-T:
\begin{equation}
\bm{z}_i^{l_1}
=
\operatorname{MM\text{-}T}^{l_1}
\left(
\bm{z}_i,
\token_{\text{img}},
\token_{\text{alb}}^i,
\token_{\text{mat}}
\right).
\label{eq:z_fusion_mr}
\end{equation}
The fused view latents are then processed by the same MV-T formulation in Sec.~\ref{sec:mv-dit}.

We optimize the MR generation model with a flow-matching loss $\mathcal{L}_{\text{flow}}^{\text{MR}}$ and the photometric representation alignment loss:
\begin{equation}
\mathcal{L}_{\text{total}}^{\text{MR}}
=
\mathcal{L}_{\text{flow}}^{\text{MR}}
+
\lambda_{\text{R}}
\mathcal{L}_{\mathrm{REPA}},
\label{eq:l_mr}
\end{equation}
where $\lambda_{\text{R}}$ balances the representation alignment term.

\section{Experiments}

We compare \method~against open-source and commercial state-of-the-art PBR generation methods through qualitative comparisons, quantitative metrics, and a user study in Sec.~\ref{sec:comparison}, comprehensively examining both intrinsic PBR accuracy and human visual preference. We also ablate main factors of our architecture in Sec.~\ref{sec:ablation}.

\subsection{Implementation Details}
\subsubsection{Dataset.}
We curate a dataset of 130K objects from Objaverse-XL~\citep{deitke2023objaverse,deitke2023objaversexl} and TexVerse~\cite{zhang2025texverse}.
For diversity, each object is rendered into 32 reference images under random cameras and environment maps. For supervision, we extract ground-truth PBR maps (albedo, metallic, roughness), normals (geometry, shading), and render photometric images across six $1024^2$ orthogonal views. We provide more details in the supplementary.

\subsubsection{Training.}

We use the intrinsic material token adapter to integrate the Qwen3.5-4B VLM~\cite{qwen3.5} with our diffusion framework.
The adapter is a two-layer MLP with channel dimensions $\{12288, 4096, 3072\}$ and an intermediate SiLU activation.
Hierarchical visual features are extracted from intermediate VLM layers $\mathcal{M}=\{8,16,24\}$.
Both the albedo and MR models initialize their DiT backbones from a pretrained text-to-image diffusion model.
The multi-view models are trained at a per-view resolution of $768^2$.
Each model is trained on 32 NVIDIA A100 GPUs for about 2 days.

\subsubsection{Inference.}
At inference time, we first generate six-view albedo maps, which then condition the MR network to estimate corresponding metallic and roughness maps. All maps per view are upscaled to $1536^2$ by a super-resolution module and mapped onto 4K UV textures. To ensure complete surface coverage, unobserved regions not captured by the orthogonal projections are inpainted using the novel view synthesis model from LumiTex~\cite{lumitex2026} (see original paper for details).

\definecolor{tabfirst}{HTML}{CEDBEE}
\definecolor{tabsecond}{HTML}{E2E9F5}
\definecolor{tabthird}{HTML}{F0F4FA}

\begin{table}[t]
\centering
\fontsize{7.5}{9}\selectfont
\caption{\textbf{Quantitative comparison with open-source PBR texture generation methods}.
All textures are evaluated at resolution $1024^2$ for a fair comparison.
``M'' and ``R'' denote ``metallic'' and ``roughness'', respectively.}
\vspace{-1em}
\setlength{\tabcolsep}{4pt}
\resizebox{\linewidth}{!}{
\begin{tabular}{lccccccc}
\toprule
Method
& {FID$\downarrow$}
& {CLIP-FID$\downarrow$}
& {CMMD$\downarrow$}
& {CLIP-I$\uparrow$}
& {LPIPS$\downarrow$}
& {MSE$_{\text{M}}\downarrow$}
& {MSE$_{\text{R}}\downarrow$}\\
\midrule

Paint-it
& 270.7 & 31.65 & 2.694 & 0.8809 & 0.2514 & 57.64 & 41.43 \\

DreamMat
& 272.3 & 30.90 & 2.592 & 0.8868 & 0.2464 & 107.4 & 39.84 \\

Hunyuan3D-2.1
& 189.2
& 18.33
& 1.758
& 0.9302
& 0.1585
& \cellcolor{tabsecond}50.04
& 32.73 \\

LumiTex
& 159.5
& \cellcolor{tabthird}14.98
& 1.431
& \cellcolor{tabthird}0.9439
& \cellcolor{tabthird}0.1286
& 54.32
& \cellcolor{tabthird}25.96 \\

{TRELLIS.2}
& \cellcolor{tabthird}{158.7}
& {16.29}
& \cellcolor{tabthird}{1.351}
& {0.9382}
& {0.1295}
& \cellcolor{tabthird}{50.54}
& {28.23} \\

{Ours (w/o joint)}
& \cellcolor{tabsecond}{145.3}
& \cellcolor{tabsecond}{14.32}
& \cellcolor{tabsecond}{1.188}
& \cellcolor{tabsecond}{0.9454}
& \cellcolor{tabsecond}{0.1198}
& {53.03}
& \cellcolor{tabfirst}{\bfseries 16.49} \\

Ours
& \cellcolor{tabfirst}\bfseries{144.3}
& \cellcolor{tabfirst}\bfseries{12.92}
& \cellcolor{tabfirst}\bfseries{1.081}
& \cellcolor{tabfirst}\bfseries{0.9510}
& \cellcolor{tabfirst}\bfseries{0.1106}
& \cellcolor{tabfirst}\bfseries{46.77}
& \cellcolor{tabsecond}16.57 \\

\bottomrule
\end{tabular}
}
\label{tab:comparisons}
\vspace{-.5em}
\end{table}

\definecolor{tabfirst}{HTML}{CEDBEE}
\definecolor{tabsecond}{HTML}{E2E9F5}
\definecolor{tabthird}{HTML}{F0F4FA}

\begin{table}[t]
\centering
\fontsize{7.5}{9}\selectfont
\vspace{-0.2em}
\caption{
\textbf{User study results.} Comparison of rendering, albedo quality, and PBR material accuracy against the latest commercial methods.
}
\vspace{-1em}
\setlength{\tabcolsep}{2pt}
\resizebox{0.96\linewidth}{!}{
\begin{tabular}{lccccc}
\toprule
Method & Average$\uparrow$ & Rendering$\uparrow$ & Albedo$\uparrow$ & Metallic$\uparrow$ & Rough.$\uparrow$ \\
\midrule
Hyper3D v2.0~\cite{HYPER_AI}
& 2.19 & 1.97 & 1.93 & 2.39 & 2.47   \\
Hunyuan3D v3.1~\cite{lai2025hunyuan3d25highfidelity3d}
& 2.85 & 3.01 & 3.01 & 2.85 & 2.55   \\
TripoAI v3.1 (4K)~\cite{TRIPO_AI}
& 2.95 & \cellcolor{tabthird}3.17 & 3.03 & \cellcolor{tabthird}3.04 & 2.53   \\
Meshy 6 (HD)~\cite{MESHY_AI}
& \cellcolor{tabthird}3.13
& \cellcolor{tabsecond}3.36
& \cellcolor{tabsecond}3.41
& 2.96
& \cellcolor{tabthird}2.83   \\
Seed3D 2.0~\cite{SEED3D}
& \cellcolor{tabsecond}3.20
& 2.96
& \cellcolor{tabthird}3.29
& \cellcolor{tabsecond}3.36
& \cellcolor{tabsecond}3.17 \\
Ours
& \cellcolor{tabfirst}\bfseries 4.31
& \cellcolor{tabfirst}\bfseries 4.55
& \cellcolor{tabfirst}\bfseries 4.33
& \cellcolor{tabfirst}\bfseries 4.21
& \cellcolor{tabfirst}\bfseries 4.11 \\
\bottomrule
\end{tabular}
}
\label{tab:userstudy}
\vspace{-.5em}
\end{table}

\begin{figure}[t]
    \centering
    \includegraphics[width=0.9\linewidth]{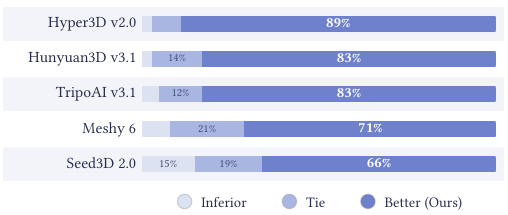}
    \vspace{-1.2em}
    \caption{\textbf{The result of the user study}. We provide the pairwise comparison result with the latest commercial methods.}
    \label{fig:fig-pairwise}
    \vspace{-1.2em}
\end{figure}

\begin{figure*}[ht]
    \centering
    \begin{minipage}{\linewidth}
        \centering
        \includegraphics[width=0.92\linewidth]{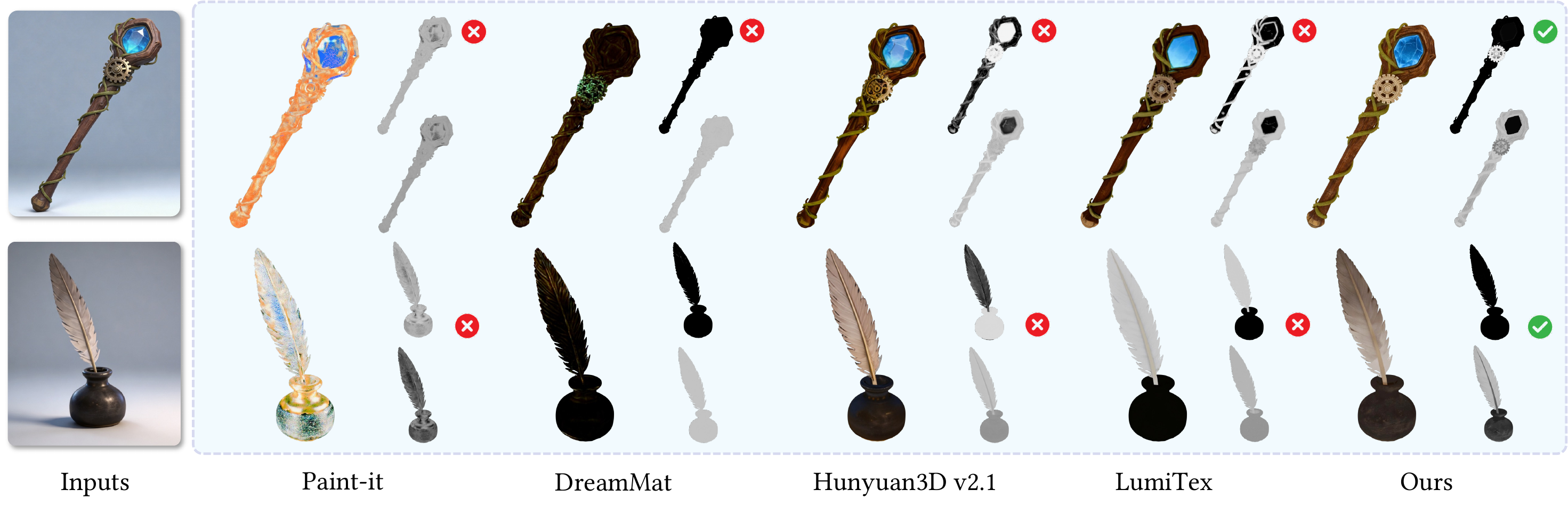}
        \vspace{-1.2em}
        \caption{\textbf{Comparison with open source methods}. Our method effectively eliminates baked-in lighting and shadows, ensuring the generation of physically accurate material attributes.
        We show the albedo channel (left) alongside the metallic (top right) and roughness (bottom right) channels.
        }
        \vspace{.6em}
        \label{fig:comp-open}
    \end{minipage}
    \begin{minipage}{\linewidth}
        \centering
        \includegraphics[width=0.92\linewidth]{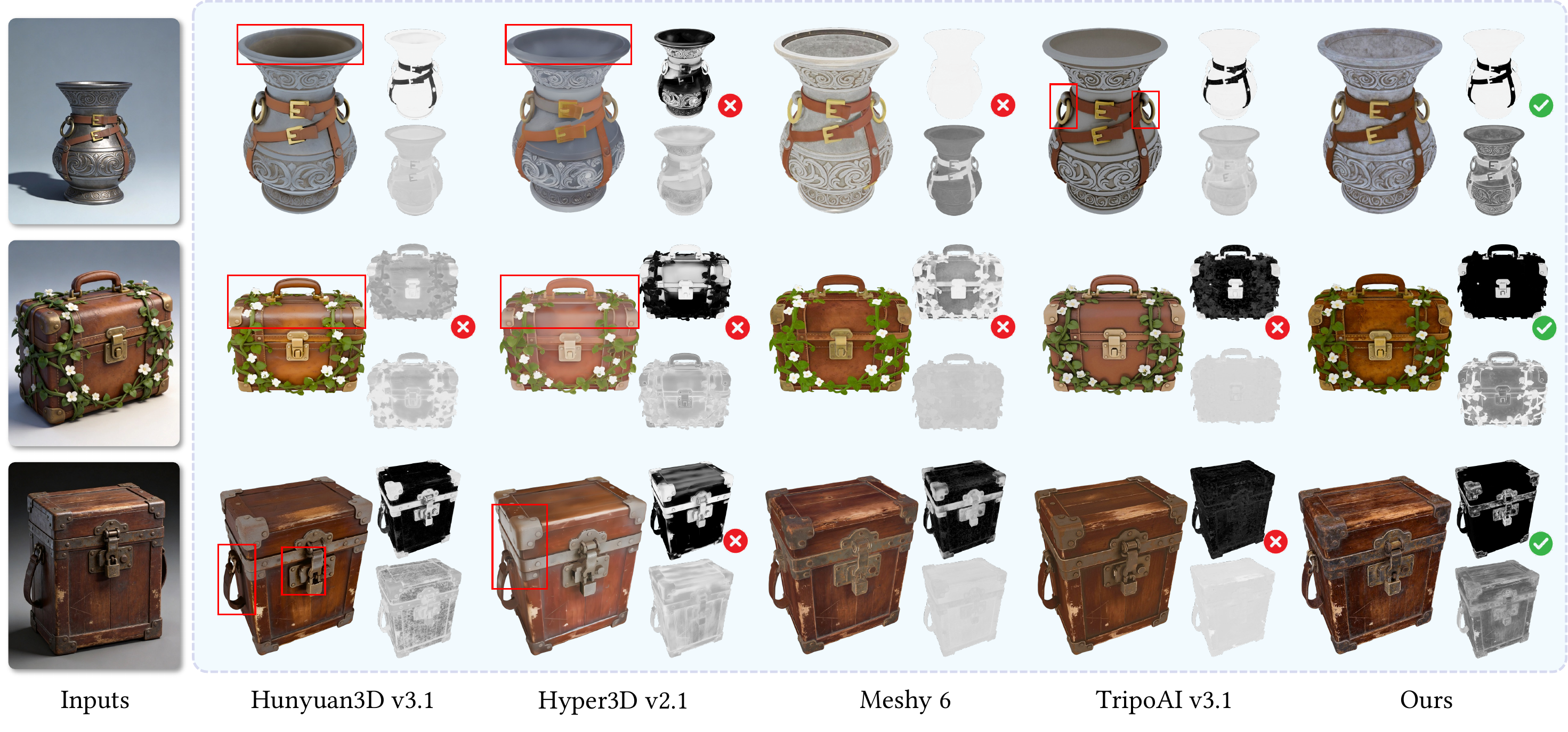}
        \vspace{-1.2em}
        \caption{\textbf{Comparison with SoTA commercial methods.} Recent commercial methods often struggle with accurate metallic and roughness estimation. By integrating VLM-based physical reasoning and photometric consistency, UltraTex generates more accurate material attributes and richer material details. For each asset, we show the albedo channel (left) alongside the metallic (top right) and roughness (bottom right) channels.}
        \label{fig:comp-commercial}
    \end{minipage}
    \vspace{-0.6em}
\end{figure*}

\subsection{Comparison with Existing Methods}
\label{sec:comparison}

\begin{figure}
    \centering
    \includegraphics[width=\linewidth]{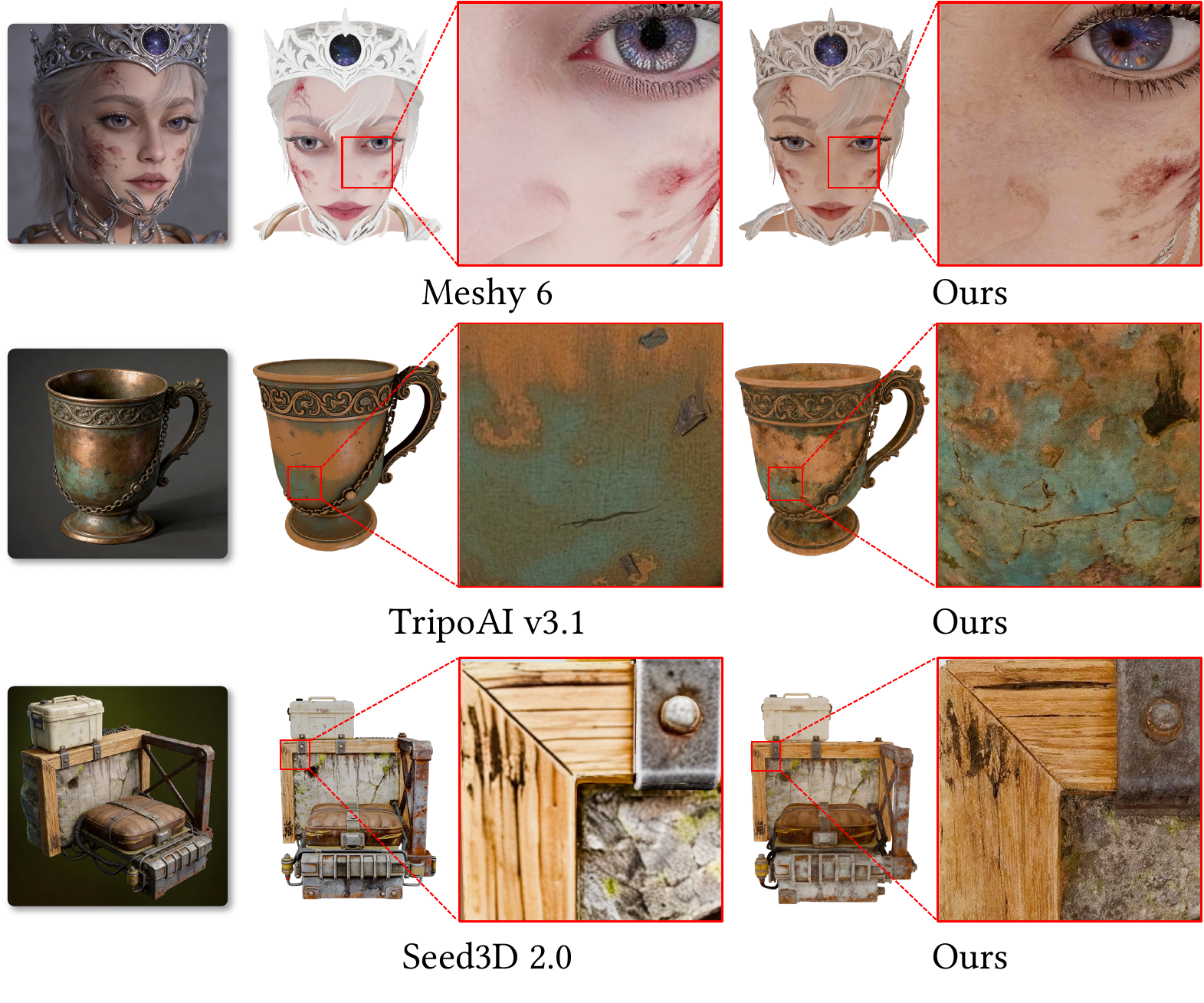}
    \vspace{-2.5em}
    \caption{\textbf{Comparison of local details.} Our model excels in generating ultra-fine albedo details with high fidelity, even for challenging surfaces such as oxidized metals and those characterized by intricate micro-textures.}
    \label{fig:local_details}
    \vspace{-1.2em}
\end{figure}

\begin{figure}[t]
    \centering
    \includegraphics[width=\linewidth]{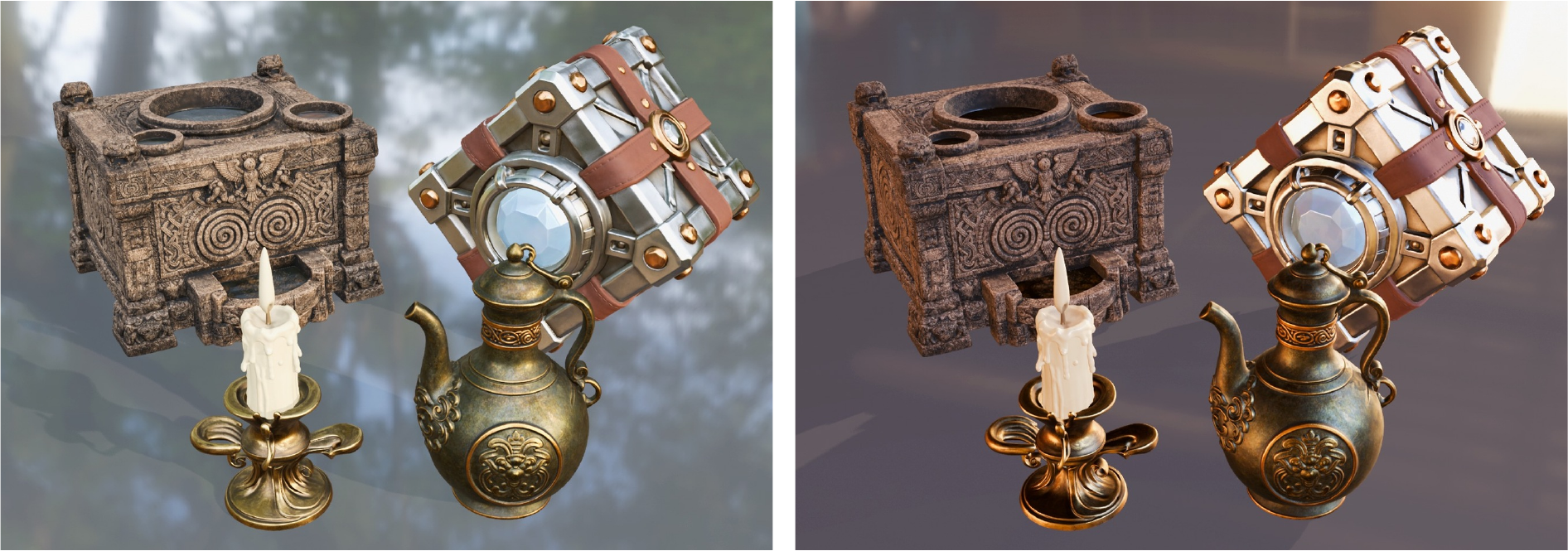}
    \vspace{-2em}
    \caption{\textbf{Relighting Results.} Our generated PBR textures exhibit physically plausible and consistent material responses under novel illumination conditions, faithfully reflecting changes in highlights and reflections.}
    \label{fig:relight}
    \vspace{-1em}
\end{figure}

\begin{figure}[t]
    \centering
    \includegraphics[width=0.92\linewidth]{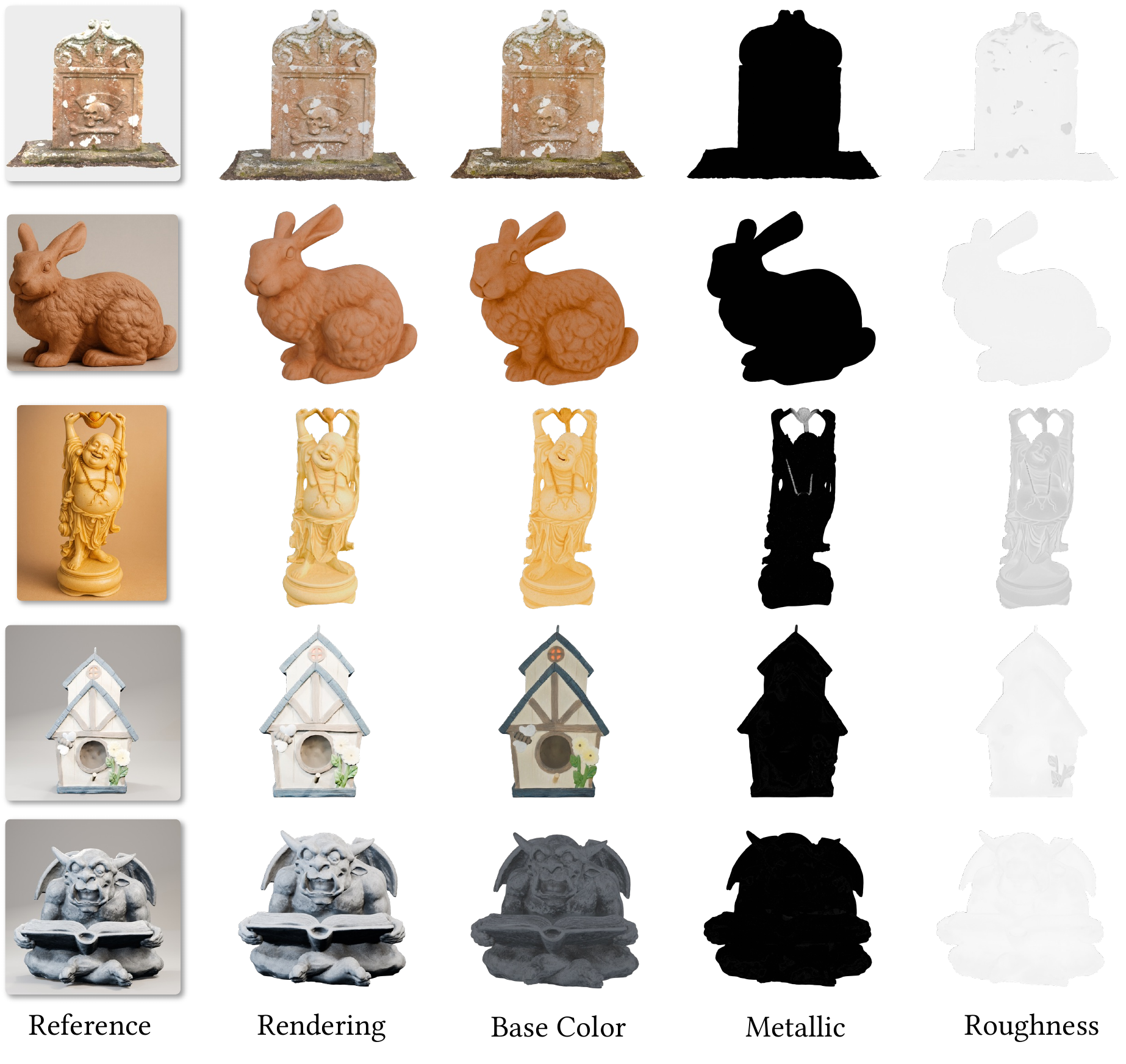}
    \vspace{-1.5em}
    \caption{\textbf{Results on real-world captures.} Our method produces accurate and physically plausible PBR textures from casually captured images and scanned meshes, demonstrating robustness to real-world inputs.}
    \vspace{-1em}
    \label{fig:real_world}
\end{figure}

\begin{figure*}[t]
    \centering
    \includegraphics[width=0.9\linewidth]{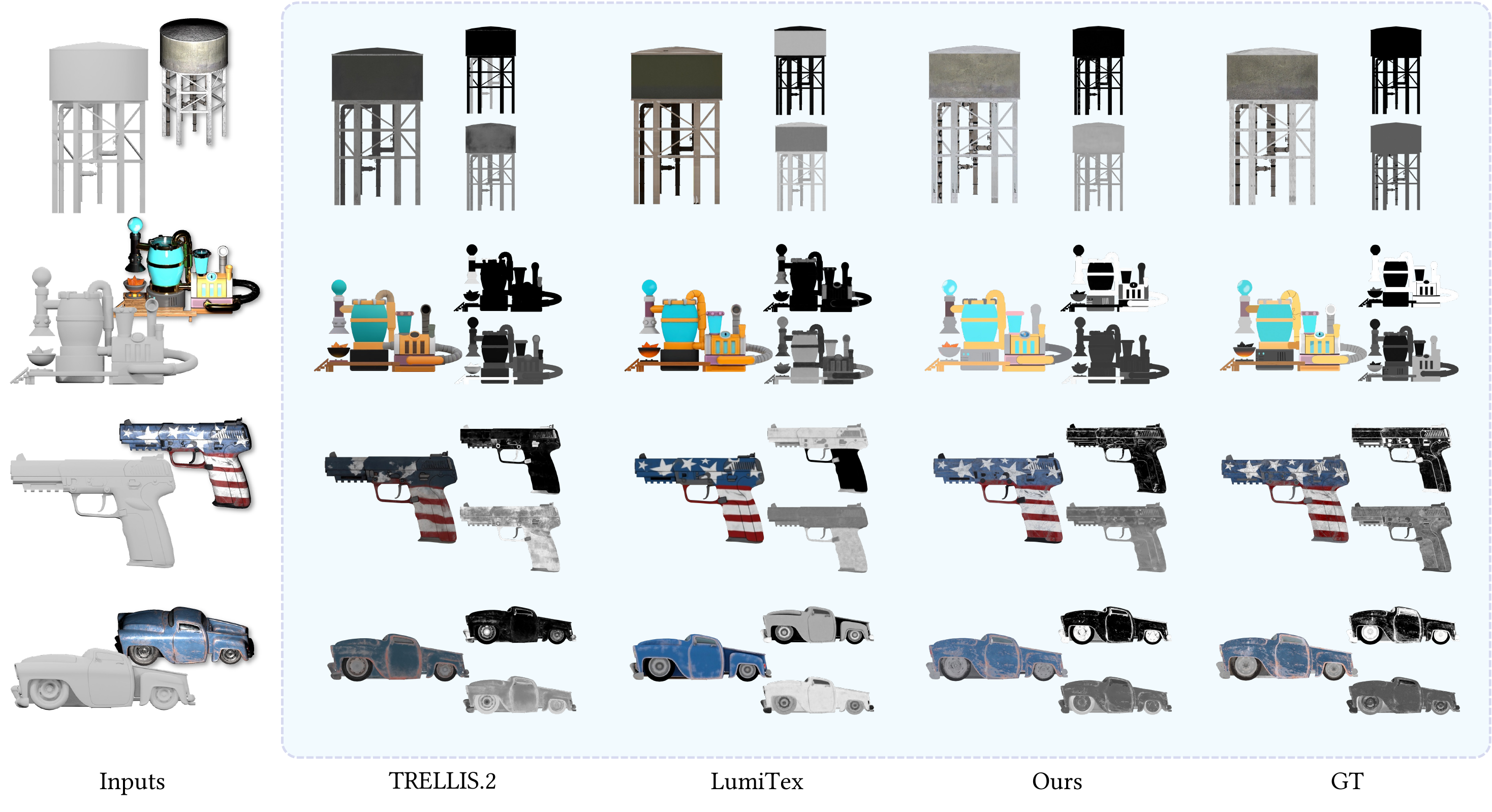}
    \vspace{-1.5em}
    \caption{\textbf{Qualitative comparison with ground-truth PBR maps}. Given the same clay mesh and reference image, our method more faithfully recovers the ground-truth material decomposition than the state-of-the-art multi-view and 3D-native baselines, with finer base-color details, cleaner metallic separation, and more accurate spatially varying roughness. Each object has: the albedo on the left, the metallic on the top right, and the roughness on the bottom right.}
    \label{fig:comp_gt}
\end{figure*}

\subsubsection{Baselines.}

We compare \method~with representative open-source PBR texture synthesis methods, including Paint-it~\citep{youwang2024paintit}, DreamMat~\citep{zhang2024dreammat}, Hunyuan3D v2.1 \citep{hunyuan3d2025hunyuan3d2.1, materialanything}, LumiTex~\citep{lumitex2026}, {and TRELLIS.2~\citep{xiang2025trellis2}.}
We also compare against commercial methods, including Hunyuan3D v3.1, Hyper3D~\citep{HYPER_AI}, Meshy 6~\citep{MESHY_AI}, and TripoAI v3.1~\citep{TRIPO_AI} and Seed3D 2.0~\cite{SEED3D}.

\subsubsection{Evaluation Metrics}
To evaluate texture fidelity, we follow the protocol of Hunyuan3D~\citep{hunyuan3d2025hunyuan3d2.1,lai2025hunyuan3d25highfidelity3d,zhao2025hunyuan3d} by employing FID~\citep{heusel2017fid}, CLIP-FID, and LPIPS~\citep{zhang2018lpips}. For a more fine-grained assessment, we use CLIP Maximum-Mean Discrepancy (CMMD)~\citep{jayasumana2024cmmd} to measure detail diversity and CLIP-I~\citep{radford2021learning} for prompt alignment. Notably, all aforementioned perceptual and semantic metrics are computed exclusively on the albedo maps. For MR estimation, accuracy is validated via the MSE ($\times 10^{-3}$). We also conduct a user study to benchmark texture quality based on human preferences.

\begin{table}[t]
\centering
\fontsize{7.5}{9}\selectfont
\caption{\textbf{Quantitative comparison of environment relighting results}. We evaluate 152 objects across four views under sampled environment maps and compare the relighting results against ground-truth renderings.}
\vspace{-1em}
\setlength{\tabcolsep}{6pt}
\begin{tabular}{lccccc}
\toprule
Method
& FID$\downarrow$
& CLIP-FID$\downarrow$
& CMMD$\downarrow$
& CLIP-I$\uparrow$
& LPIPS$\downarrow$ \\
\midrule
TRELLIS.2
& 158.4
& 22.89
& 1.8896
& 0.9049
& 0.1213 \\

LumiTex
& 144.4
& 17.87
& 1.6943
& 0.9249
& 0.1110 \\

Ours
& \textbf{111.2}
& \textbf{15.08}
& \textbf{1.3575}
& \textbf{0.9370}
& \textbf{0.0988} \\
\bottomrule
\end{tabular}

\label{tab:relighting}
\vspace{-1.5em}
\end{table}

\subsubsection{Quantitative Results}

We quantitatively compare \method{} with baselines on 152 held-out Objaverse-XL objects, using Blender-rendered GT assets as reference inputs and rendering outputs at $1024^2$ resolution.
Except for MSE on MR attributes, all perceptual and semantic metrics are computed on rendered albedo/MR images, covering visual fidelity, semantic alignment, and PBR accuracy.

As shown in Tab.~\ref{tab:comparisons}, our method consistently outperforms existing baselines across all metrics.
It achieves better FID and LPIPS for perceptual quality, lower CMMD for texture diversity, higher CLIP-I/CLIP-FID for semantic alignment, and the lowest MSE across PBR attributes, validating its superior physical accuracy.
{
Additionally, Table~\ref{tab:relighting} further shows that our method outperforms both the state-of-the-art multi-view baseline (LumiTex) and the 3D-native baseline (TRELLIS.2) under environment relighting.
}

To evaluate perceptual quality, we conduct a blind user study where 31 professional 3D modelers rate results from 1 to 5 based on overall rendering, albedo, roughness, and metallic fidelity.
\method{} achieves the highest scores across all dimensions (Tab.~\ref{tab:userstudy}) and substantially outperforms competitors in pairwise win rates (Fig.~\ref{fig:fig-pairwise}), demonstrating strong expert preference.

\subsubsection{Qualitative Results}
We qualitatively compare \method~with recent open-source and commercial baselines for PBR texture generation. For test data, we generate reference images via Seedream 4.5~\cite{seedream2025seedream}, followed by mesh generation with Hunyuan3D v3.1, then apply each method for texture generation.

As shown in Fig.~\ref{fig:comp-open} and \ref{fig:comp-commercial}, existing methods like Hunyuan3D and Hyper3D often suffer from illumination leakage, producing albedo maps with baked-in lighting artifacts. Furthermore, most baselines have difficulty resolving intrinsic material ambiguity, frequently misclassifying metallic regions or producing over-smoothed roughness maps.
In contrast, by leveraging material knowledge and illumination cues, \method~consistently produces more accurate metallic estimates and fine-grained roughness details, while also delivering albedo maps with superior physical fidelity and structural richness.

This advantage is especially evident in challenging cases (Fig.~\ref{fig:local_details}), including micro-scale roughness variations, oxidized or coated metals, and part-wise material changes, where baselines often fail. Furthermore, the diverse gallery in Fig.~\ref{fig:supp_more} demonstrates our model’s ability to produce physically plausible textures across a wide array of geometric complexities and material compositions.

{We further evaluate the generated PBR textures through relighting under diverse novel illumination conditions. As shown in Fig.~\ref{fig:relight}, the recovered material properties respond naturally to changes in lighting, producing consistent highlights, reflections, and surface appearance across different environments.}

{To further evaluate the accuracy of intrinsic material decomposition, we compare against baselines on objects with ground-truth PBR maps. Given the same clay mesh and reference image, \method~more faithfully recovers the target material properties than the multi-view baseline LumiTex and the 3D-native baseline TRELLIS.2, particularly in base color, metallic separation, and spatially varying roughness (Fig.~\ref{fig:comp_gt}).}

{
\subsubsection{Results on Real-World Inputs}
We further evaluate \method{} on casually captured real-world inputs and scanned meshes, including three examples collected from Sketchfab and two from the DTC dataset~\citep{dtcDong_2025_CVPR}. These inputs exhibit uncontrolled illumination, appearance ambiguity, and imperfect scanned geometry. As shown in Fig.~\ref{fig:real_world}, \method{} remains robust under these conditions, producing coherent PBR materials and physically plausible relighting. The results demonstrate that our method generalizes beyond clean rendered inputs to practical real-world asset creation.
}

\subsection{Ablation Study}
\label{sec:ablation}

\begin{figure}
    \centering
    \includegraphics[width=\linewidth]{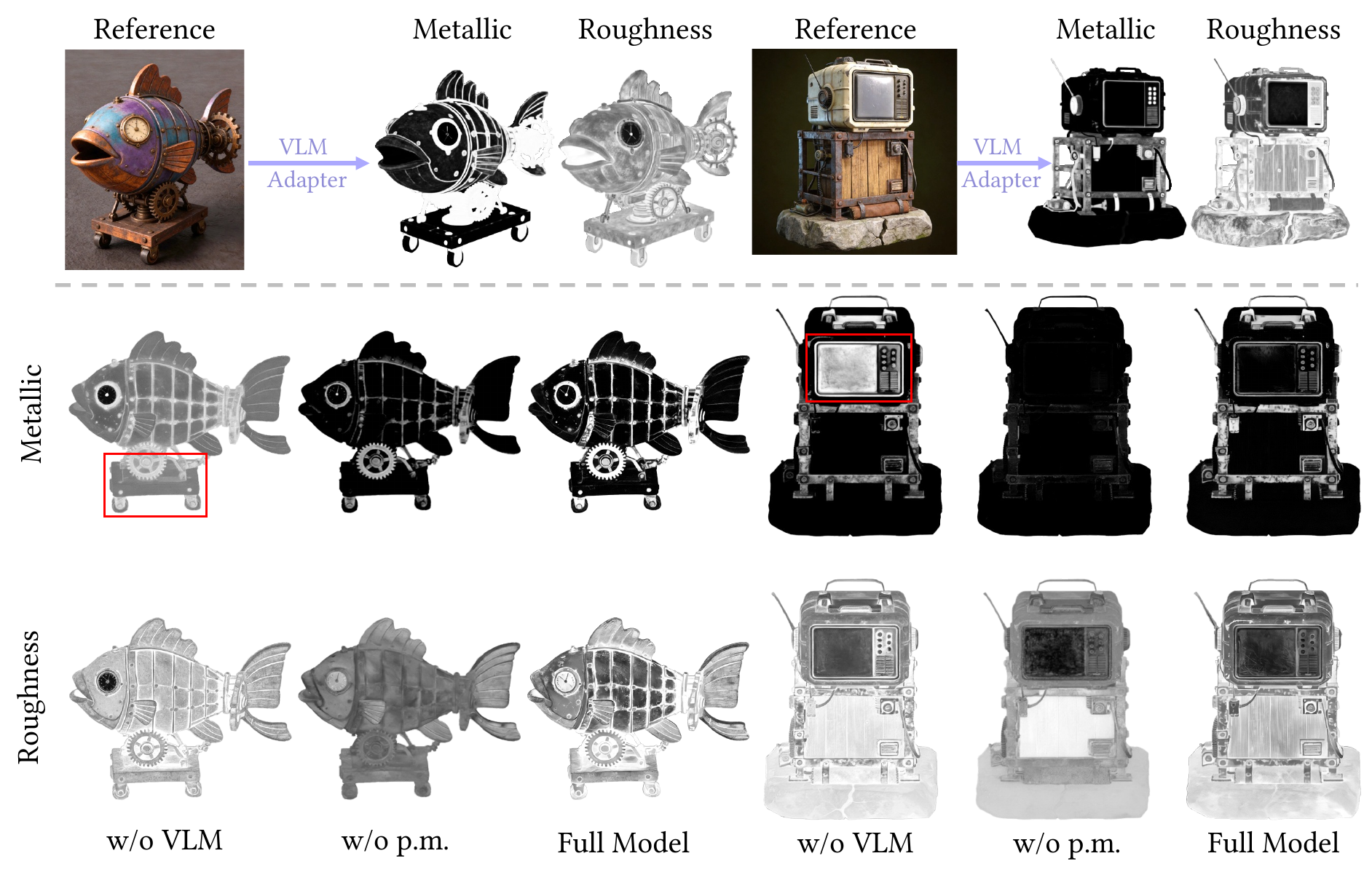}
    \vspace{-2.5em}
    \caption{\textbf{Impact of semantic and photometric alignment}. The first row visualizes our VLM adapter capturing intrinsic material semantics. The following rows compare metallic and roughness maps of ablated variants and our full model. Without VLM-derived tokens, the network incorrectly predicts non-metallic surfaces (e.g., painted fish, plastic TV casing) as metallic. Removing photometric alignment yields weaker metallic contrast and fails to recover structural details like gears, rivets, and screws.}
    \label{fig:ablation}
    \vspace{-1em}
\end{figure}

\paragraph{VLM-derived Intrinsic Material Token}
To validate the impact of VLM priors, we train a variant of MR estimation network without intrinsic material token guidance.
{
As shown in Fig.~\ref{fig:ablation} and Tab.~\ref{tab:core_ablation}, removing the VLM-derived tokens leads to clear material misclassification and increases both metallic and roughness errors.
}
The ablated model relies heavily on low-level visual cues, such as color and specular highlights, and therefore mistakes strong reflections for metallic materials.
This confirms that VLM guidance provides essential semantic and part-level material priors, enabling more physically grounded MR estimation that is robust to misleading appearance-based heuristics.
{
In addition, removing the visual conditions also degrades MR quality, demonstrating their complementary role in providing dense spatial guidance for material generation.
}

{
\paragraph{Decoding Probe of VLM Encoder}
To quantify the intrinsic material information encoded by the 1D material tokens, we conduct a decoding probe by replacing the VLM encoder with VAE and DINOv2 encoders under the same setting and directly decoding MR maps from the resulting tokens.
As shown in Tab.~\ref{tab:decoding_probe}, VLM features consistently outperform VAE and DINOv2 features, demonstrating that the resulting tokens carry richer material semantics and better disentangle intrinsic MR properties from observed appearance.
This is because intrinsic material properties require semantic reasoning beyond observable appearance, which is better captured by the world knowledge encoded in VLM representations.
}

\begin{figure}[t]
    \centering
    \includegraphics[width=\linewidth]{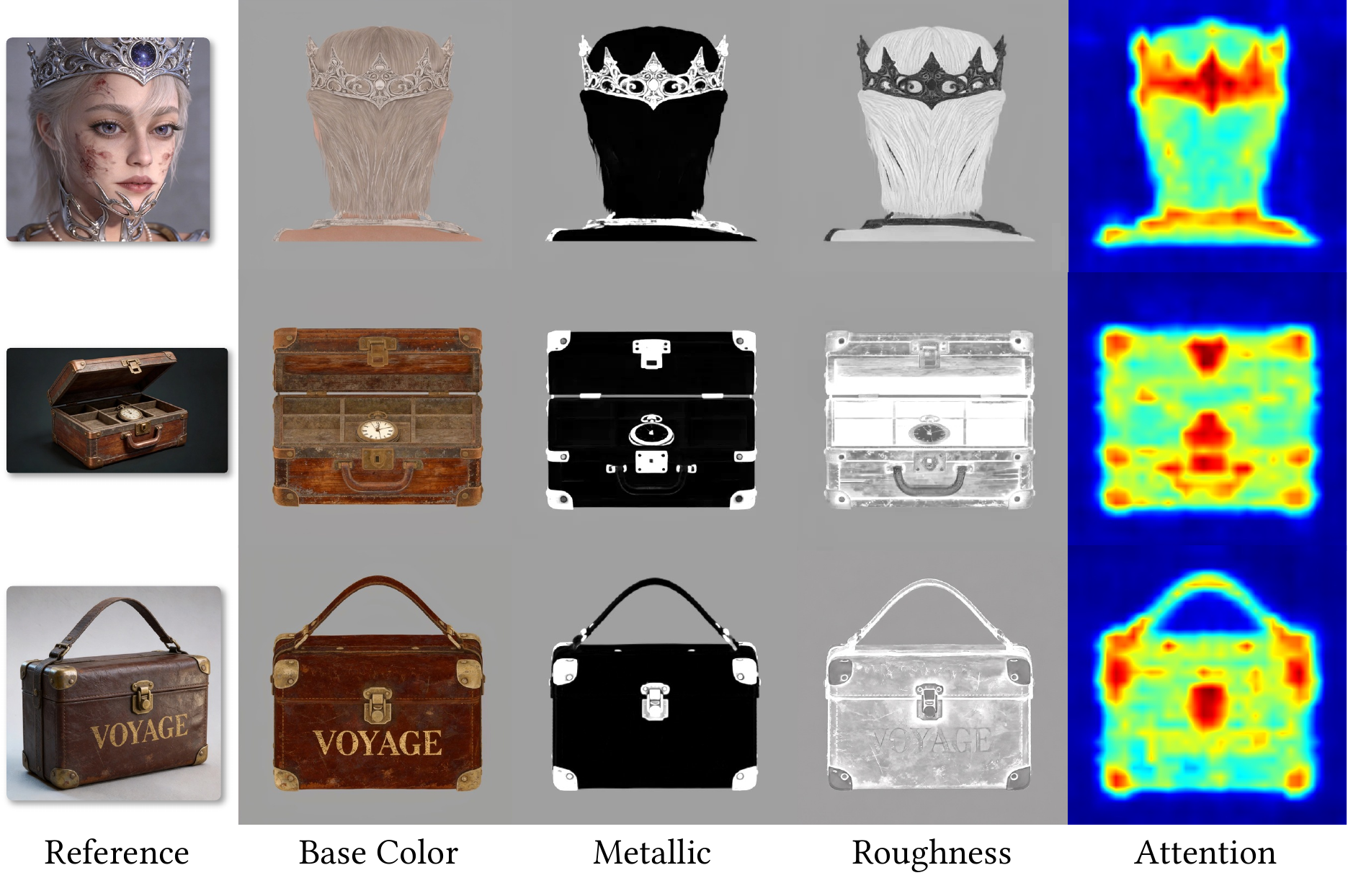}
    \vspace{-2em}
    \caption{\textbf{VLM-token attention visualization}. We visualize the attention map from VLM material tokens to noisy image tokens. The attention maps exhibit clear spatial localization over material-relevant object parts, such as metallic ornaments and hardware, indicating that the 1D material tokens capture part-aware material correspondence.}
    \vspace{-1.5em}
    \label{fig:attention}
\end{figure}

{
\paragraph{VLM-token Attention Map Visualization}
To better understand the part-level spatial localization of the 1D material tokens, we visualize the attention from VLM tokens to noisy image tokens at the 4th layer of the Multi-Modal DiT. As shown in Fig.~13, the responses are strongly localized to material-relevant object parts, such as metallic ornaments, buckles, and other hardware regions. Notably, the VLM tokens themselves carry no explicit spatial coordinates; their spatial correspondence emerges through interaction with the image tokens inside the DiT. This indicates that the material tokens are not merely used as global semantic conditions, but establish part-level correspondence with regions exhibiting the associated material properties.
}

\begin{table}[t]
\centering
\fontsize{7.5}{9}\selectfont
\caption{\textbf{Quantitative ablation study.} We evaluate the contribution of each core component by removing visual conditions, VLM guidance, or photometric alignment.
MSE values are reported in units of $10^{-3}$.}
\vspace{-1em}
\setlength{\tabcolsep}{8pt}
\begin{tabular}{lcc}
\toprule
Method
& MSE$_{\mathrm{M}}$ $\downarrow$
& MSE$_{\mathrm{R}}$ $\downarrow$ \\
\midrule
Ours (w/o visual conditions)
& 55.46
& 28.23 \\

Ours (w/o VLM)
& 54.85
& 22.50 \\

Ours (w/o photometric alignment)
& 52.32
& 18.45 \\

Ours
& \textbf{46.77}
& \textbf{16.57} \\
\bottomrule
\end{tabular}

\label{tab:core_ablation}
\vspace{-0.5em}
\end{table}

\begin{table}[t]
\centering
\fontsize{7.5}{9}\selectfont
\caption{\textbf{Quantitative comparison of different encoder designs}.
For each encoder, we train only the adapter while keeping the image diffusion model frozen, and evaluate the resulting MR map generation.
MSE values are reported in units of $10^{-3}$. ``ORM'' denotes the channel combination of ``one'', ``roughness'' and ``metallic''.}
\vspace{-1em}
\setlength{\tabcolsep}{8pt}
\begin{tabular}{lccc}
\toprule
Encoder
& {$\mathrm{MSE}_{\mathrm{M}}$ $\downarrow$}
& {$\mathrm{MSE}_{\mathrm{R}}$ $\downarrow$}
& {LPIPS (ORM)$\downarrow$} \\
\midrule

Image VAE
& 58.04
& 34.01
& 0.4085 \\

DINOv2
& 43.76
& 32.09
& 0.3592 \\

VLM (Qwen3.5)
& \bfseries 27.93
& \bfseries 15.79
& \bfseries 0.2065 \\

\bottomrule
\end{tabular}

\label{tab:decoding_probe}
\vspace{-.5em}
\end{table}

\paragraph{Photometric Regularization Loss}
To investigate the role of representation level photometric alignment, we compare our full model against a variant trained without this mechanism. As shown in Fig.~\ref{fig:ablation}, removing this constraint causes the model to regress toward conservative averaged values, resulting in less reliable MR predictions. This degradation is especially prominent in roughness estimation, where high-frequency variations on materials such as painted metal, wood, and stone are suppressed.
These results demonstrate that intrinsic reflectance properties cannot be robustly inferred from semantic priors alone; the model requires explicit cues regarding surface response to illumination. Photometric alignment supplies essential, complementary reflectance information for physically accurate MR estimation.

\paragraph{Joint Shading Normal Estimation}
To evaluate the impact of our joint prediction strategy, we compare our model against a baseline trained solely on albedo.
As shown in Fig.~\ref{fig:bump_res}, omitting joint shading-normal estimation leads to visibly smoother albedo maps and weaker surface structures, especially around fine engravings and material boundaries.
This confirms that joint generation is essential for preserving rich, high-frequency surface details, and significantly enhances the visual realism of the final renderings through these enriched surface bump.

\begin{table}[t]
\centering
\fontsize{7.5}{9}\selectfont
\caption{\textbf{VAE reconstruction quality.} We evaluate reconstruction fidelity on 366 MR view-maps using the RGB-pretrained VAE.}
\vspace{-0.8em}
\setlength{\tabcolsep}{10pt}
\begin{tabular}{lccc}
\toprule
& PSNR$\uparrow$ & SSIM$\uparrow$ & LPIPS$\downarrow$ \\
\midrule
VAE Recon.
& 38.656
& 0.9839
& 0.0051 \\
\bottomrule
\end{tabular}

\label{tab:vae_reconstruction}
\vspace{-1.5em}
\end{table}

{
\paragraph{VAE Reconstruction Quality}
MatLat~\cite{yeo2026matlat} points out a domain gap between general-purpose image VAE latent spaces and material-property distributions. This issue is also relevant to our framework, as we exploit an image-model prior and encode metallic--roughness (MR) maps with an RGB-pretrained VAE.
We therefore compute standard VAE reconstruction metrics on 366 MR view maps by comparing the original and reconstructed MR maps.
As shown in Tab.~\ref{tab:vae_reconstruction}, although the VAE is not specifically trained for PBR material maps, our evaluation shows that it retains the key material information with minimal reconstruction distortion, supporting its use for MR generation in our framework.
}

{
\subsection{Applications}

\begin{figure}[t]
    \centering
    \includegraphics[width=\linewidth]{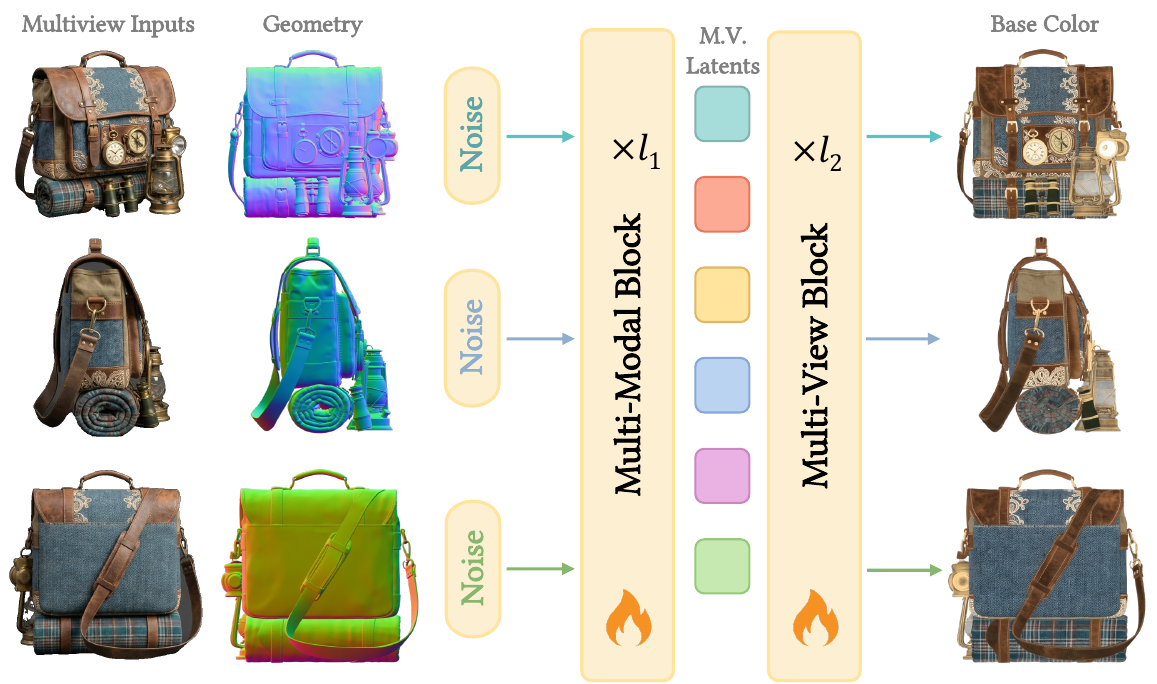}
    \vspace{-2em}
    \caption{\textbf{Multi-view input pipeline}. Our framework naturally extends to multi-view references by independently encoding each input view and pairing its image features with the corresponding geometry tokens in the Multi-Modal DiT, while leaving the remaining generation pipeline unchanged.}
    \label{fig:multiview}
    \vspace{-1em}
\end{figure}

\paragraph{Multi-view Input Extension}
Multi-view references are common in practical content-creation workflows, as they provide complementary observations of object appearance and reduce ambiguity in occluded regions and material properties. As illustrated in Fig.~\ref{fig:multiview}, our framework can be naturally extended to support multi-view inputs with only a minor modification to the Multi-Modal DiT, while leaving the remaining generation pipeline unchanged.
Specifically, given a set of reference images $\{\bm{I}_i\}_{i=1}^{N}$, we construct an image token for each view by replacing Eq.~\ref{eq:token_img} with
\begin{equation}
\token_{\mathrm{img}}^{i}
=
\mathrm{Linear}_{\mathrm{img}}
\left[
\mathcal{E}_{\mathrm{VAE}}(\bm{I}_i);
\operatorname{DINOv2}(\bm{I}_i)
\right],
\label{eq:token_img_multiview}
\end{equation}
where each reference view is independently encoded into its corresponding appearance representation. We then pair the image token with the geometry token of the same view and perform multi-modal fusion as
\begin{equation}
\bm{z}_{i}^{l_1}
=
\operatorname{MM\text{-}T}^{l_1}
\left(
\bm{z}_{i},
\token_{\mathrm{img}}^{i},
\token_{\mathrm{geo}}^{i}
\right).
\label{eq:z_fusion_multiview}
\end{equation}
This view-wise formulation preserves the original architecture and enables the model to incorporate complementary appearances from multiple observations with minimal changes to the original pipeline.

\begin{figure}[t]
    \centering
    \includegraphics[width=\linewidth]{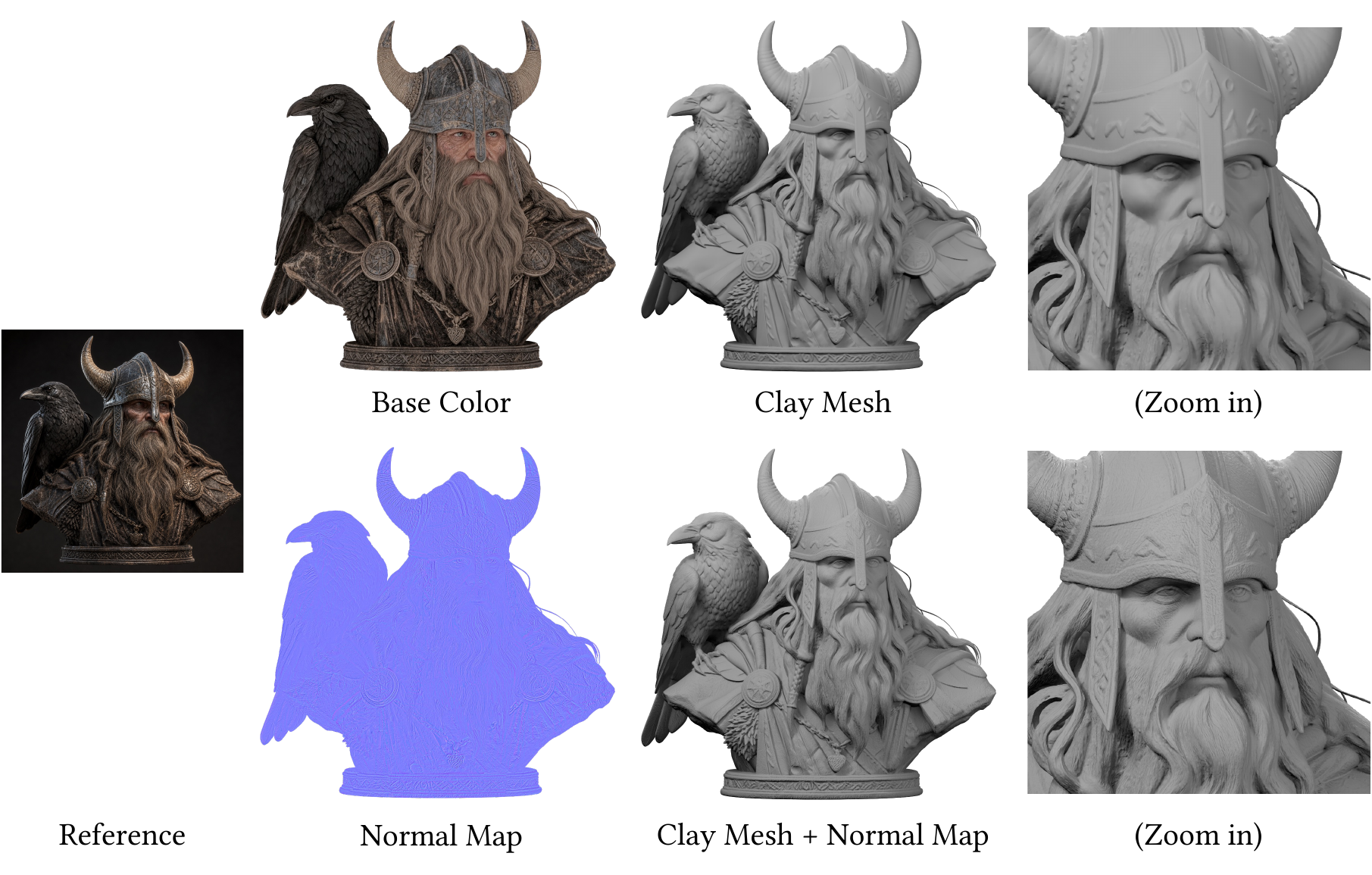}
    \vspace{-2em}
    \caption{\textbf{Normal map transformation}. Our predicted shading normals are converted to a tangent-space normal map, adding fine-scale surface details to the original clay mesh without modifying its geometry.}
    \label{fig:normal_map}
    \vspace{-1em}
\end{figure}

\paragraph{Normal Map Transformation}
Our predicted shading normals can be directly converted into a standard tangent-space normal map using the local surface frame of the mesh. As shown in Fig.~\ref{fig:normal_map}, applying the resulting normal map to the original clay mesh enriches its shading with fine-scale surface details without modifying the underlying geometry. In particular, subtle structures such as the helmet engravings, facial details, and hair strands become substantially more pronounced, producing a more detailed and realistic surface appearance.

\begin{figure}[t]
    \centering
    \includegraphics[width=\linewidth]{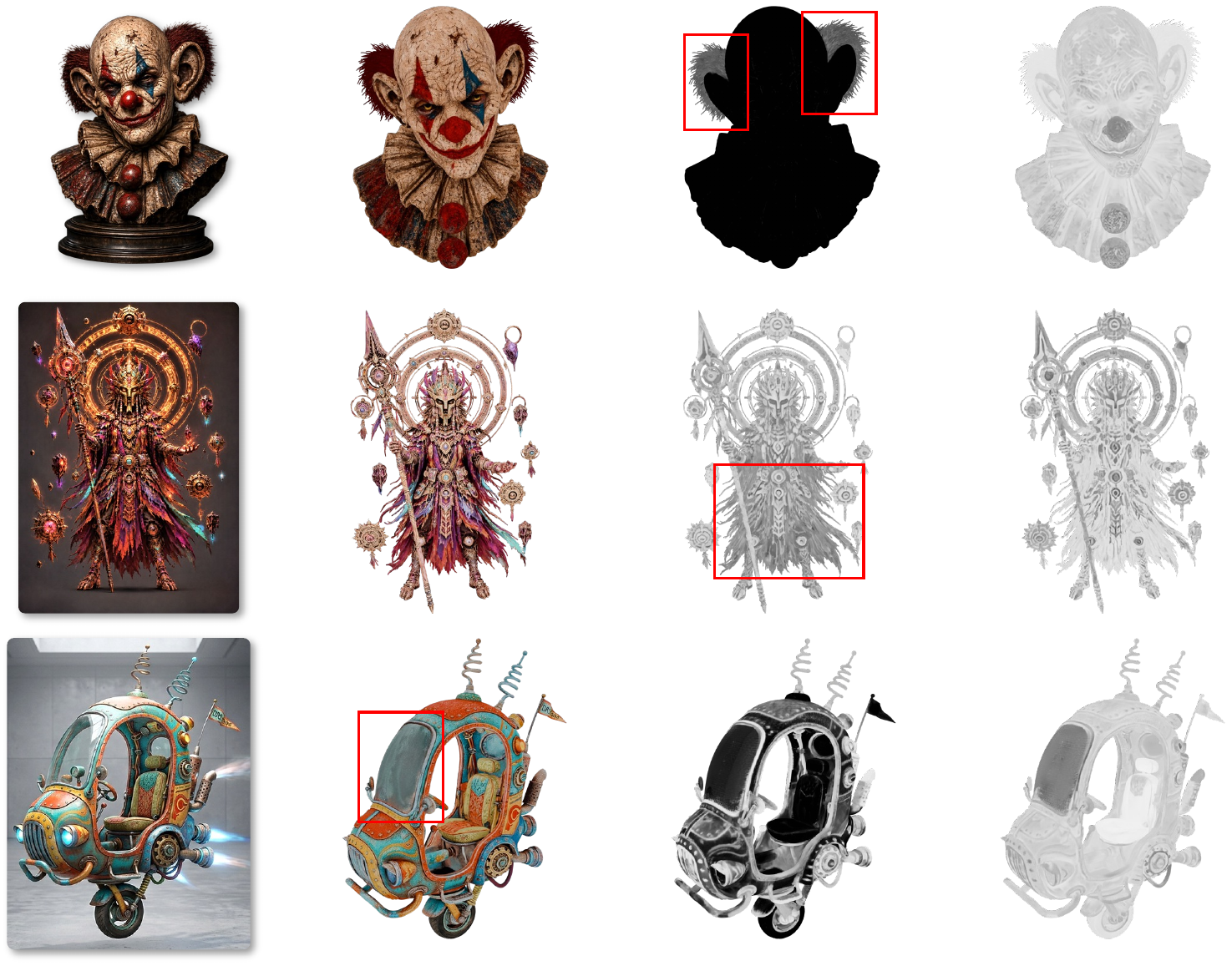}
    \vspace{-2em}
    \caption{\textbf{Failure Cases}. Highly stylized assets may use MR for artistic rather than physical effects, leading to ambiguous decomposition, as seen in the joker's hair and the character's garments. In addition, effects beyond the opaque metallic--roughness model, such as the vehicle's transparent windshield, cannot be faithfully represented.}
    \label{fig:failure}
    \vspace{-1em}
\end{figure}

\subsection{Failure Cases}
Our main failure mode arises on highly stylized UGC assets, where PBR attributes are often manipulated to achieve a desired artistic appearance rather than to faithfully represent physical material properties. In particular, metallic may serve as an appearance control for highlights, contrast, or stylized reflectance, making the decomposition fundamentally ambiguous from appearance alone. As shown in Fig.~\ref{fig:failure}, the joker's hair and the character's garments exhibit simultaneously high metallic and roughness values---an artistically plausible choice that does not correspond to their physical material identity. Our predictions may therefore remain visually and physically plausible while differing from such intentionally non-physical reference values. A separate limitation arises from material effects beyond our current PBR parameterization: for example, the transparent windshield of the stylized vehicle in Fig.~\ref{fig:failure} cannot be faithfully represented by the opaque metallic--roughness model.

}

\begin{figure*}
    \includegraphics[width=\linewidth]{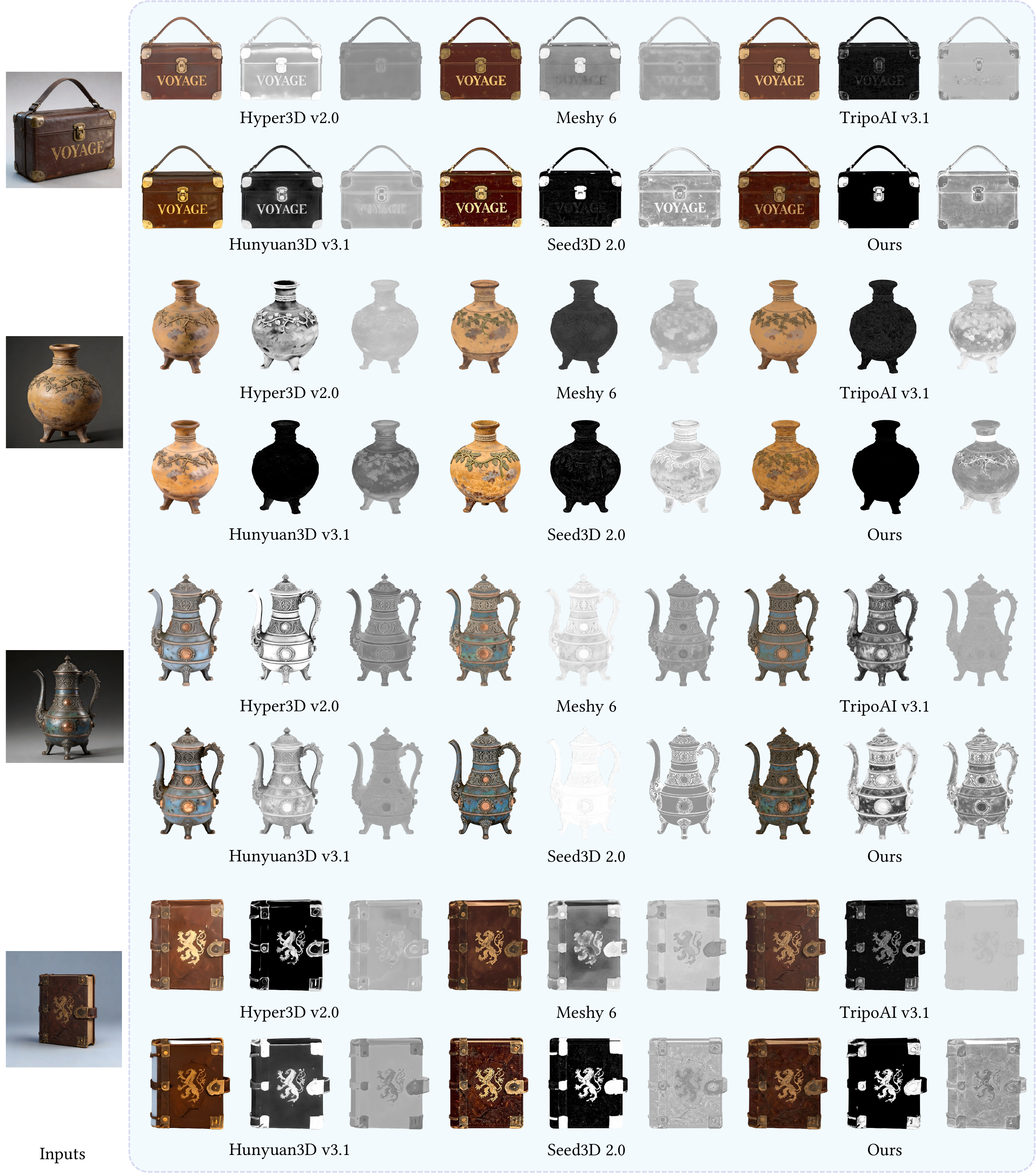}
    \caption{\textbf{More Qualitative Results.} We present additional PBR comparisons with other methods (from left to right: albedo, metallic, roughness). Zooming in for details. }
    \label{fig:supp_comp}
\end{figure*}

\begin{figure*}
    \centering
    \includegraphics[width=0.95\linewidth]{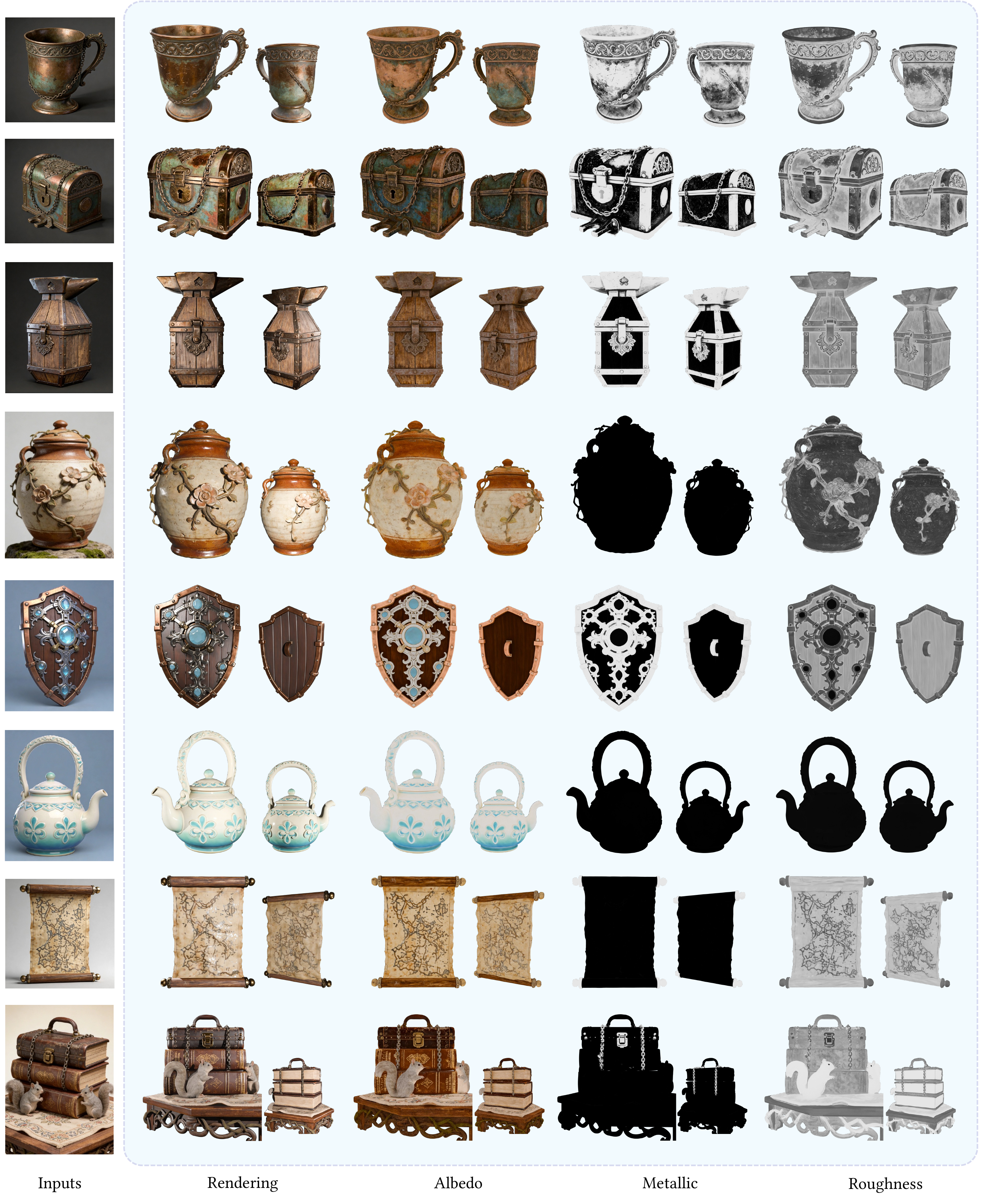}
    \vspace{-1em}
    \caption{\textbf{More results.} We provide additional multi-view renderings with their corresponding generated PBR maps. Zooming in is recommended to observe the ultra-fine albedo and material details.}
    \label{fig:supp_more}
\end{figure*}

\section{Conclusions}

We presented \method{}, a cascaded multi-view framework for single-image PBR texture generation that moves beyond appearance matching by aligning physical knowledge with diffusion-based material generation.
By integrating a VLM-guided intrinsic material adapter for part-wise semantic priors and photometric representation alignment for material-light interaction, \method{} generates high-fidelity metallic and roughness maps with improved physical plausibility and relighting behavior.
In addition, joint estimation of world-space shading normals with albedo provides geometry-aware supervision, enriching surface detail and improving albedo fidelity.
Comprehensive evaluations on diverse assets demonstrate that \method{} substantially outperforms existing open-source and commercial methods in visual fidelity, material accuracy, and physical plausibility, highlighting its potential for high-quality 3D asset creation.

\paragraph{Limitations and Future Work}
Despite its strong generative capability, \method{} still faces challenges with complex materials involving non-standard light transport.
For example, as shown in Fig.~\ref{fig:supp_more}, the model may approximate gemstones using albedo patterns and low roughness values, but cannot faithfully represent transparency, translucency, caustics, or more detailed BRDF effects.
This limitation stems partly from the restricted material channels used in current training data, which focus on albedo, metallic, and roughness.
A promising direction is to incorporate additional material channels, such as specular, coat, transmission, and subsurface scattering, once sufficiently large-scale datasets with such annotations become available.
Another direction is to extend the framework toward more expressive material models while preserving the robustness and scalability of multi-view PBR generation.

\begin{acks}
This work is partially supported by the Shenzhen Loop Area Institute under grant FPF10120260001 and Guangdong Provincial Fund for Distinguished Young Scholars No. 2023B1515020055.
\end{acks}

\bibliographystyle{ACM-Reference-Format}
\bibliography{sample-bibliography}

\end{document}